\documentclass[pdflatex,sn-mathphys-num]{sn-jnl}

\usepackage{graphicx}%
\usepackage{multirow}%
\usepackage{amsmath,amssymb,amsfonts}%
\usepackage{amsthm}%
\usepackage{mathrsfs}%
\usepackage[title]{appendix}%
\usepackage{textcomp}%
\usepackage{manyfoot}%
\usepackage{booktabs}%
\usepackage{algorithm}%
\usepackage{algorithmicx}%
\usepackage{algpseudocode}%
\usepackage{listings}%
\usepackage{tikz}
\usetikzlibrary{arrows}

\usepackage{tabularx}
\usepackage{array}
\usepackage{longtable}
\usepackage{caption}
\usepackage{graphicx}

\usepackage[colorinlistoftodos]{todonotes}

\theoremstyle{thmstyleone}%
\theoremstyle{thmstyletwo}%

\theoremstyle{thmstylethree}%

\begin{document}

\title{Differential privacy for medical deep learning: methods, tradeoffs, and deployment implications}

\author[1]{\fnm{Marziyeh} \sur{Mohammadi}}
\author[1]{\fnm{Mohsen} \sur{Vejdanihemmat}}
\author[2]{\fnm{Mahshad} \sur{Lotfinia}}
\author[3,4]{\fnm{Mirabela} \sur{Rusu}}
\author[2]{\fnm{Daniel} \sur{Truhn}}
\author[1]{\fnm{Andreas} \sur{Maier}}
\author*[1,2,3,4]{\fnm{Soroosh} \sur{Tayebi Arasteh}}\email{soroosh.arasteh@rwth-aachen.de}
\affil[1]{\orgname{Pattern Recognition Lab, Friedrich-Alexander-Universität Erlangen-Nürnberg}, \orgaddress{\city{Erlangen}, \country{Germany}}}
\affil[2]{\orgname{Department of Diagnostic and Interventional Radiology, University Hospital RWTH Aachen}, \orgaddress{\city{Aachen}, \country{Germany}}}
\affil[3]{\orgname{Department of Radiology, Stanford University}, \orgaddress{\city{Stanford, CA}, \country{USA}}}
\affil[4]{\orgname{Department of Urology, Stanford University}, \orgaddress{\city{Stanford, CA}, \country{USA}}}

\abstract{
Differential privacy (DP) is a key technique for protecting sensitive patient data in medical deep learning (DL). As clinical models grow more data-dependent, balancing privacy with utility and fairness has become a critical challenge. This scoping review synthesizes recent developments in applying DP to medical DL, with a particular focus on DP-SGD and alternative mechanisms across centralized and federated settings. Using a structured search strategy, we identified 74 studies published up to March 2025. Our analysis spans diverse data modalities, training setups, and downstream tasks, and highlights the tradeoffs between privacy guarantees, model accuracy, and subgroup fairness. We find that while DP, especially at strong privacy budgets ($\epsilon \approx $10), can preserve performance in well-structured imaging tasks, severe degradation often occurs under strict privacy ($\epsilon \approx $1), particularly in underrepresented or complex modalities. Furthermore, privacy-induced performance gaps disproportionately affect demographic subgroups, with fairness impacts varying by data type and task. A small subset of studies explicitly addresses these tradeoffs through subgroup analysis or fairness metrics, but most omit them entirely. Beyond DP-SGD, emerging approaches leverage alternative mechanisms, generative models, and hybrid federated designs, though reporting remains inconsistent. We conclude by outlining key gaps in fairness auditing, standardization, and evaluation protocols, offering guidance for future work toward equitable and clinically robust privacy-preserving DL systems in medicine.}

\maketitle


\section*{Introduction}

Deep learning (DL) has become a cornerstone of modern medical artificial intelligence (AI), driving advances in diagnostic imaging, survival prediction, biosignal analysis, and electronic health record (EHR) modeling \cite{rajpurkar2022ai, tayebi2024large, haug2023artificial, tayebi2024treasure}. However, the deployment of DL models in clinical settings raises critical concerns regarding patient privacy \cite{66kaissis2021end}. Medical data are inherently sensitive, and DL models trained on such data are vulnerable to various privacy attacks, including membership inference, model inversion, and identity disclosure, potentially exposing the presence or characteristics of individual patients in the training set~\cite{22tayebi2024preserving, usynin2021adversarial, wang2021variational, haim2022reconstructing, carlini2023extracting}.

\begin{figure}
    \centering
\includegraphics[width=\textwidth,height=\textheight,keepaspectratio]{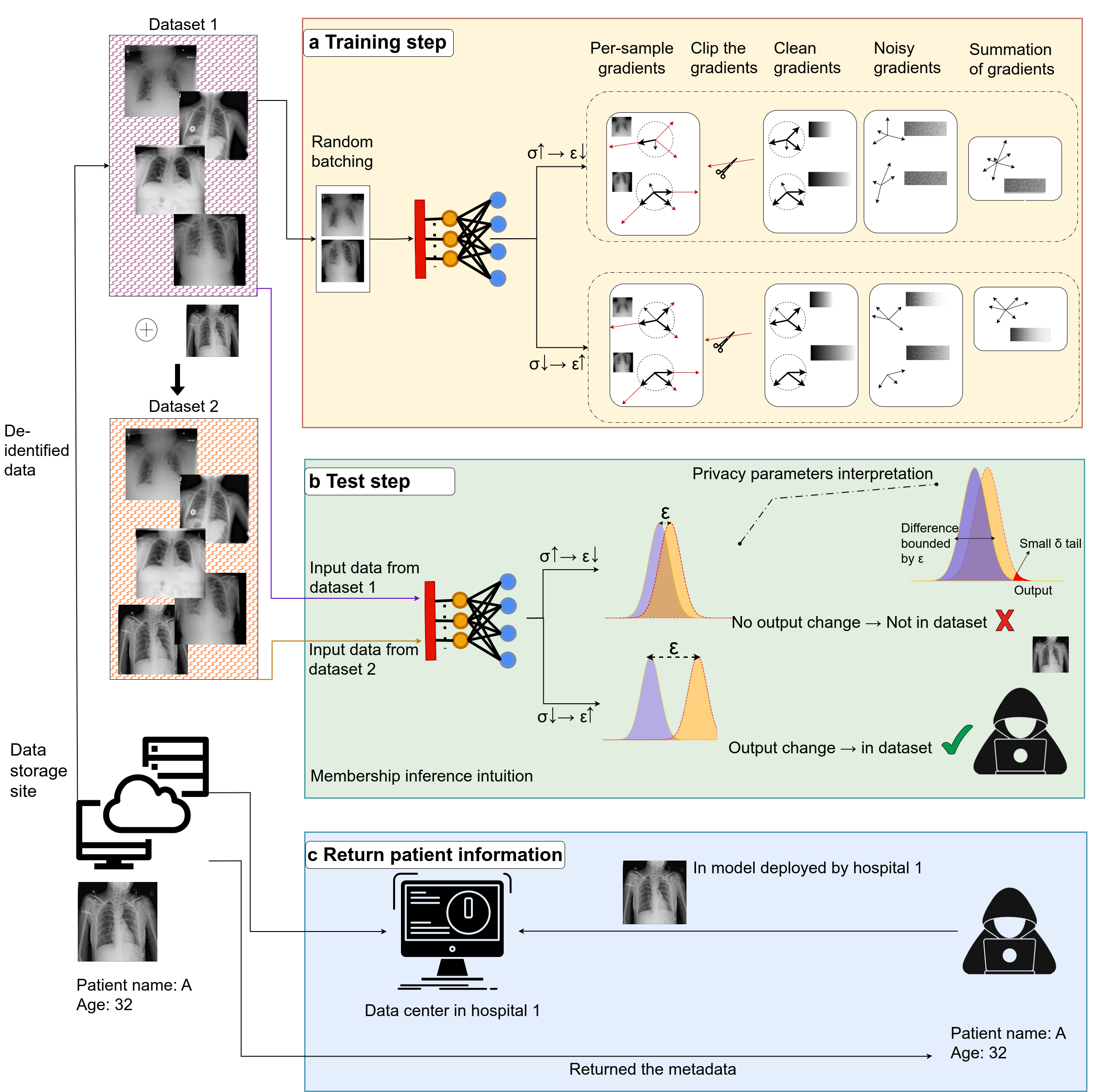}
\caption{Overview of differential privacy (DP) applied to model training and inference.
\textbf{(a)} Training samples are processed individually to compute per-sample gradients, which are clipped to limit sensitivity and then perturbed by noise sampled from a Gaussian distribution with variance $\sigma^2$. Higher noise (larger $\sigma$) yields stronger privacy (smaller $\epsilon$), and the resulting noisy gradients are aggregated to update model parameters.
\textbf{(b)} DP bounds how much the model’s output distribution can change when a single individual’s data is added or removed from the training set. When noise is low (larger $\epsilon$), outputs for neighboring datasets may differ noticeably, making membership inference easier. With stronger privacy (smaller $\epsilon$), output distributions overlap and individual influence becomes indistinguishable. The parameter $\delta$ represents the probability that the DP guarantee may not hold.
\textbf{(c)} In the absence of DP, an attacker with access to a deployed model may exploit output differences to infer whether a specific sample was included in training, potentially revealing sensitive patient information (e.g., metadata or identifiers). Representative chest X-ray images are provided by the ChestX-ray14 dataset from NIH Clinical Center \cite{wang2017chestx}.}
\label{fig:overview}
\end{figure}

Differential privacy (DP)~\cite{dwork2014algorithmic} provides formal guarantees to protect individuals within a dataset by bounding the influence of any single record on the model's output. In DL, this is commonly implemented via differentially private stochastic gradient descent (DP-SGD)~\cite{DPSGD}, which injects noise into clipped gradients during training (Fig. \ref{fig:overview}). Despite its theoretical strength, practical deployment of DP in healthcare DL introduces key tradeoffs: noise reduces model utility and may disproportionately affect underrepresented subgroups, potentially reinforcing existing healthcare inequities~\cite{tran2021decision, 13TayebiDomainTransfer, 22tayebi2024preserving}.

To better understand the state of DP in clinical DL, we conduct a comprehensive scoping review of empirical studies that apply DP mechanisms in medical contexts. While several prior reviews, such as in \cite{zhao2019differential, demelius2025recent, fu2024differentially, liu2024survey}, address DP broadly or focus on applications in generic domains, no existing synthesis systematically evaluates how DP interacts with clinical data, model design, privacy-utility tradeoffs, fairness impacts, or adversarial robustness.

To guide our review, we investigate the following research questions:
\begin{itemize}
    \item[\textbf{RQ1:}] To what extent do medical DL models leak sensitive information, and under what conditions are privacy breaches most likely?
    \item[\textbf{RQ2:}] How effectively do DP-based methods mitigate membership inference and other privacy attacks in healthcare applications?
    \item[\textbf{RQ3:}] What are the empirical impacts of DP on model performance, fairness, and clinical applicability in real-world scenarios?
\end{itemize}

This scoping review aims to critically synthesize current knowledge, identify key challenges and opportunities, and provide practical guidance for future research and clinical practice involving privacy-preserving DL models in medicine.
Throughout this review, "training" refers to model optimization on available data; "validation" refers to performance evaluation on a held-out subset of the same dataset; and "external validation" denotes evaluation on data from a different site, institution, or patient population. "Deployment" refers to releasing a model for operational clinical use (e.g., inside a hospital system or federated network). "Monitoring" refers to post-deployment performance tracking, including detection of performance degradation or subgroup fairness drift over time. In federated learning (FL) \cite{FederatedLearningSource}, "clients" denote data-holding sites; "centralized" refers to training on pooled data. We use “DP-SGD,” “local DP,” “privacy budget,” and “privacy accountant” with their standard meanings in DP literature.
In the following sections, we first review the applications of DP across clinical DL tasks, covering data modalities, training settings, and privacy accounting methods. We then analyze how architectural choices, normalization strategies, and pretraining affect the privacy-utility tradeoff, with a focus on model performance across varying privacy budgets. Next, we examine fairness impacts, highlighting subgroup disparities, the metrics used to evaluate them, and gaps in current practices. We also survey alternative privacy mechanisms beyond DP-SGD, including FL, local perturbation, and hybrid methods. Finally, we synthesize empirical findings on privacy attacks and defenses, before concluding with a discussion of key limitations, open challenges, and recommendations for building robust, equitable, and privacy-preserving AI systems in medicine.

\section*{Results}

\begin{figure}[ht]
    \centering
        \includegraphics[width= 0.98\linewidth]{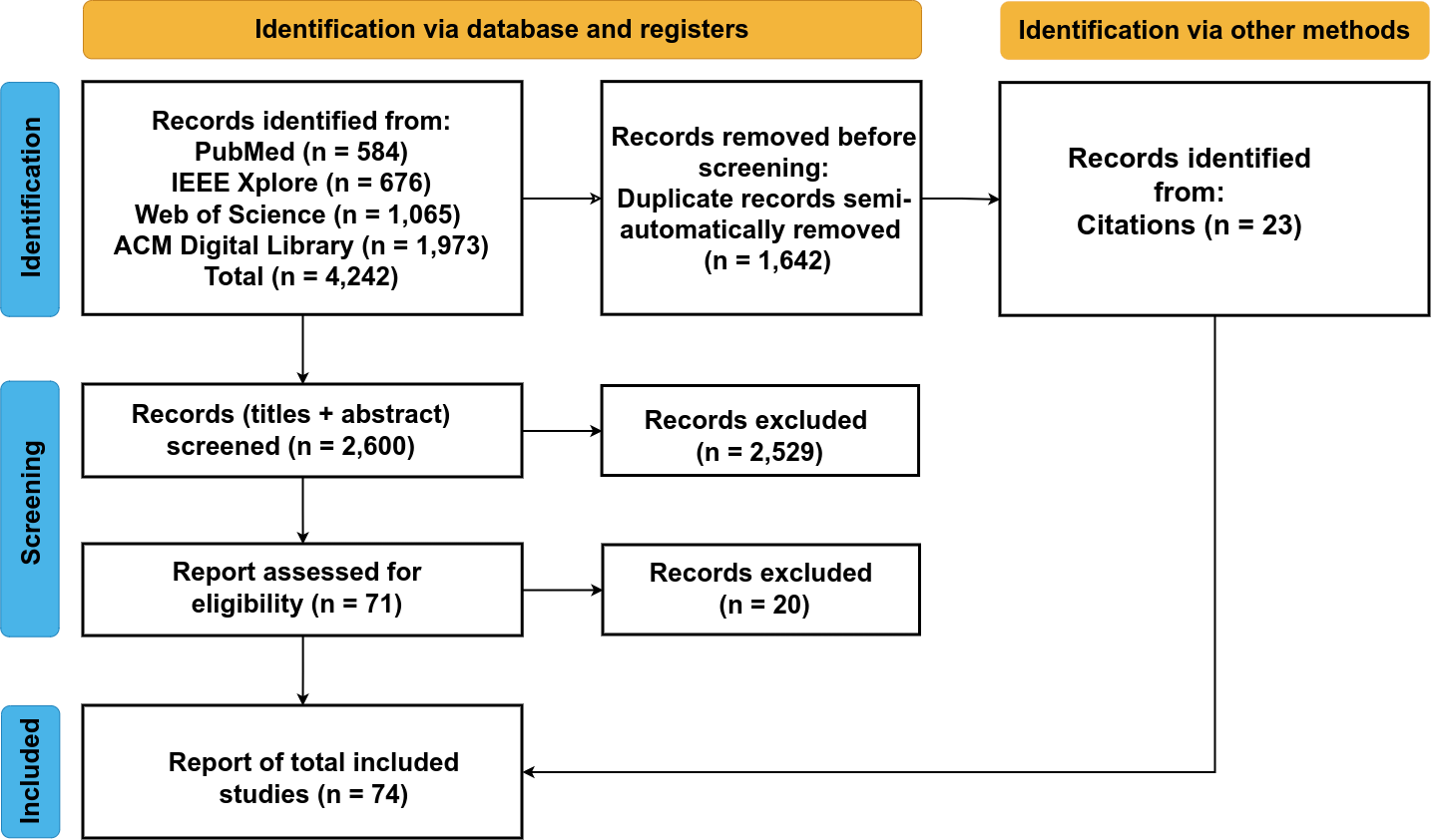}
    \caption{Preferred reporting items for systematic reviews and meta-analyses extension for scoping reviews (PRISMA-ScR) flow diagram illustrating the selection process for this study. The diagram \cite{tricco2018prisma, page2021prisma} details the number of records identified through database and manual searches, duplicates removed, records screened by title and abstract, full-text articles assessed for eligibility, and the final number of studies included.}
    \label{fig:flowdiagram}
\end{figure}

\begin{figure}
    \centering
    \includegraphics[width=\textwidth,height=\textheight,keepaspectratio]{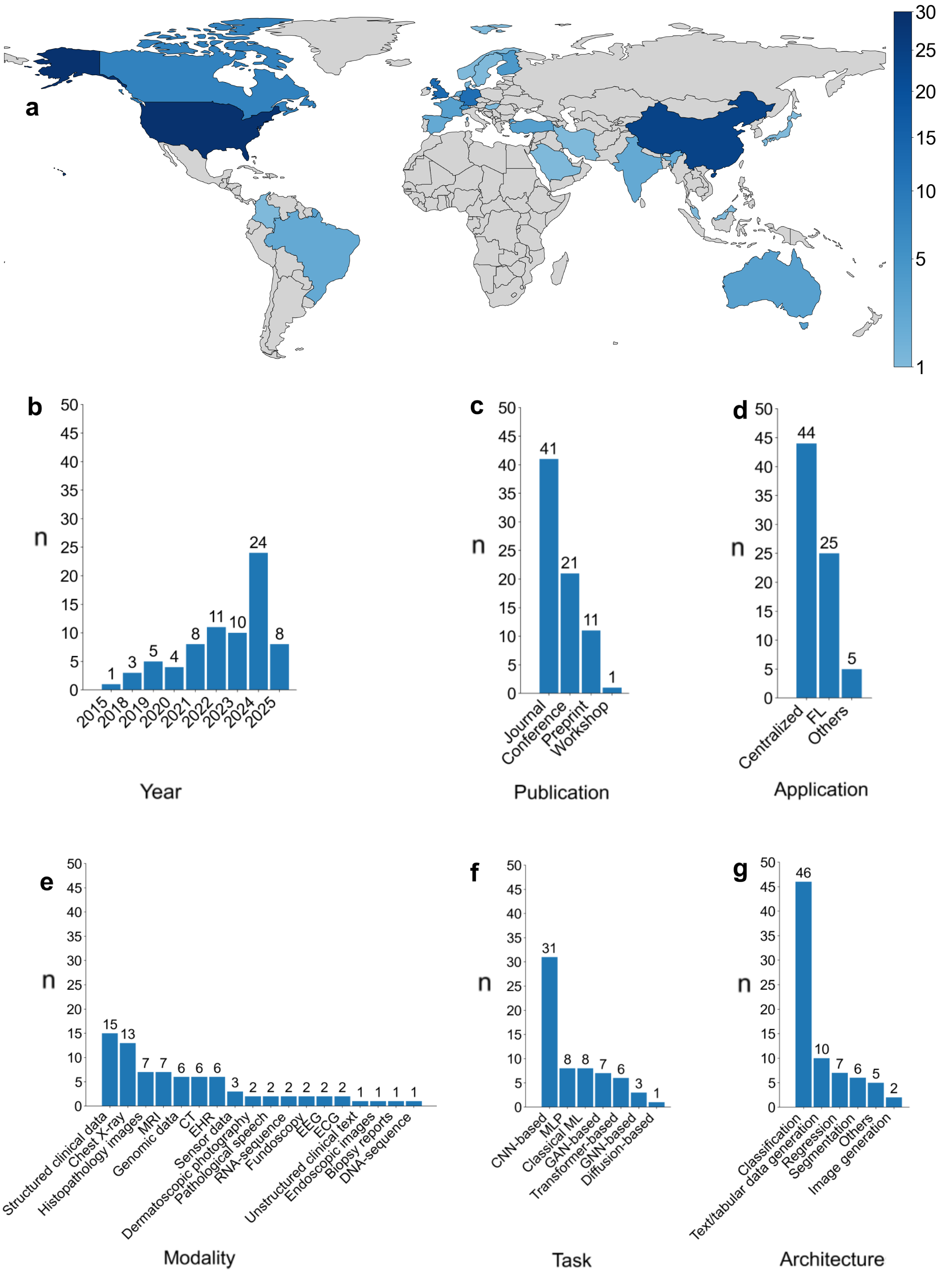}
  \centering
  \caption{Overview of study characteristics across included papers. \textbf{(a)} Geographic distribution of included studies based on author affiliations. Color intensity reflects the number of publications with at least one affiliated author per country; all affiliations of all the coauthors of each study are considered in this statistic. The top 4 contributing countries include the USA, China, the UK, and Germany, with 29, 24, 13, and 12 studies, respectively. Distributions of included studies (n=74) are shown with bar diagrams by \textbf{(b)} publication year, \textbf{(c)} publication type, and \textbf{(d)} training paradigm. Distributions of DP-SGD studies (n=67) are shown with bar diagrams by \textbf{(e)} data modality, \textbf{(f)} downstream tasks, and \textbf{(g)} architecture type. Some studies used multiple modalities, downstream tasks, or architectures; all such instances were counted, so categories are not mutually exclusive. Bar heights indicate counts. FL: federated learning; MLP: multilayer perceptron; CNN: convolution neural networks; GAN: generative adversarial network; GNN: graph neural network; MLP: multilayer perceptron; ViT: vision transformer.
    }
\label{fig:stats_further}
\end{figure}

A total of 4,242 records were retrieved through systematic searches across PubMed, IEEE Xplore, ACM Digital Library, and Web of Science. After automated and manual duplicate removal, 2,600 unique records remained. During title and abstract screening, 2,529 records were excluded based on predefined eligibility criteria, primarily due to lack of medical relevance or absence of differential privacy components. The remaining 71 articles underwent full-text assessment, where 20 additional papers were excluded due to insufficient empirical content or misalignment with the review scope. Ultimately, 51 studies were retained from the database search, and an additional 23 studies were identified through citation tracking and preprints, resulting in a total of 74 included papers. The full screening and selection workflow is shown in Fig.~\ref{fig:flowdiagram}, with descriptive publication trends summarized in Fig.~\ref{fig:stats_further}.

To structure the analysis, the included studies were first organized according to their primary privacy mechanism. Applications of DP-SGD, the most widely used approach in the field \cite{66kaissis2021end, 22tayebi2024preserving, 67ziller2024reconciling, de2022unlocking, chua2024private}, are summarized in Table~\ref{tab:dpssgd_app}. Design factors influencing the privacy–utility tradeoff, including model architectures, normalization strategies, and pretraining choices, are presented in Table~\ref{tab:utility_design}, while performance trends across different privacy budgets are compiled in Table~\ref{tab:utility_results}. Fairness evaluations and subgroup analyses are captured in Table~\ref{tab:fairness}. Studies using alternative privacy mechanisms beyond DP-SGD are listed in Table~\ref{tab:additional}, and evaluations of attack resilience are shown in Table~\ref{tab:attacks}.


\subsection*{Theoretical background}

Before presenting the empirical findings, we provide the necessary theoretical background to clarify how DP is mathematically defined and how these guarantees extend to DL workflows. This section explains the formal privacy parameters $(\epsilon, \delta)$, their composition when data or models are accessed multiple times, and how DP is operationalized in DL through gradient clipping and noise addition.

DP was introduced by Dwork et al.~\cite{TheAlgorithmicFoundationsofDifferentialPrivacy} as a mathematically rigorous framework to limit the influence of any single individual's data on the output of a computation. This protection is achieved by injecting controlled noise into a computation's output, thereby masking the presence or absence of individual records.

Consider two neighboring datasets $D$ and $D'$, differing by a single entry. A randomized algorithm $M$ is said to satisfy $(\varepsilon, \delta)$-DP if for all possible outputs $O$,
\begin{equation}\label{fr:dpdefinition}
    \Pr[M(D) \in O] \leq e^{\varepsilon} \cdot \Pr[M(D') \in O] + \delta
\end{equation}
Here, $\varepsilon$ controls the worst-case multiplicative difference in output probabilities between $D$ and $D'$, while $\delta$ allows for a small additive probability of failure. When $\delta = 0$, we recover the original, stricter definition of $\varepsilon$-DP.
In practice, $\delta$ is often set inversely proportional to dataset size ($\delta \sim 1/n$), ensuring that the likelihood of privacy failure remains negligible. Together, the $(\varepsilon,\delta)$ pair quantifies the privacy guarantee, with smaller values implying stronger protection.

Rényi differential privacy (RDP)~\cite{RDP} is a generalization of DP that leverages Rényi divergence~\cite{RD} to provide tighter composition guarantees. The Rényi divergence of order $\alpha > 1$ between the output distributions of $M$ on $D$ and $D'$ is defined as:
\begin{equation}
    D_\alpha(M(D) \| M(D')) = \frac{1}{\alpha - 1} \log \mathbb{E}_{o \sim M(D')} \left[ \left( \frac{P_{M(D)}(o)}{P_{M(D')}(o)} \right)^\alpha \right]
\end{equation}
A mechanism satisfies $(\alpha, \varepsilon)$-RDP if this divergence is bounded by $\varepsilon$ for all adjacent $D$ and $D'$:
\begin{equation}\label{renaydp}
    D_\alpha\big(M(D) \,\|\, M(D')\big) \leq \varepsilon
\end{equation}
Smaller values of $\alpha$ make the analysis more sensitive to small distributional differences, while larger values emphasize worst-case behavior. RDP can be converted to $(\varepsilon, \delta)$-DP using analytical bounds, and is widely used for accounting in iterative training.

In real-world deployments, differentially private mechanisms are rarely applied only once \cite{67ziller2024reconciling, kaissis2020secure}. Instead, the same dataset may be accessed repeatedly, for example, during iterative model training (such as DP-SGD, which performs many gradient updates), or after deployment through user queries. DP therefore requires tracking the cumulative privacy loss across all uses of the data \cite{DPSGD}.
The sequential composition property states that if two differentially private mechanisms are applied to the same dataset, with privacy parameters $(\varepsilon_1, \delta_1)$ and
$(\varepsilon_2, \delta_2)$, then releasing the outputs of both mechanisms provides
$(\varepsilon_1 + \varepsilon_2, \delta_1 + \delta_2)$-DP. Thus, each access to the data consumes part of the overall privacy budget, and the guarantee becomes weaker as more outputs are released \cite{dwork2014algorithmic}.

For iterative methods such as DP-SGD, naïve linear composition can substantially overestimate the privacy cost. RDP enables tighter accounting of cumulative privacy loss across many iterations, resulting in improved bounds such as:
\begin{equation}\label{eq:composition}
    \varepsilon_{\text{total}} \approx \sqrt{T} \cdot \varepsilon_{\text{per iteration}}
\end{equation}
where $T$ is the number of training iterations. Modern DP libraries, including Opacus \cite{opacus_neurips2021}, implement privacy accountants based on this formulation and stop training once a predefined privacy budget is exhausted.

Repeated use also introduces additional considerations once a private model is deployed. Multiple users may query the same model, and each query consumes privacy budget from the same underlying dataset. The DP guarantee assumes that users act independently; if users share outputs or coordinate queries, their combined set of queries must be treated as a single adversary, and privacy loss accumulates accordingly. In practice, complete non-collusion cannot be guaranteed, especially in clinical environments where multiple analysts or systems may interact with the same model. Therefore, deployed DP systems incorporate operational safeguards, such as authentication, per-user privacy budget allocation, query logging, audit trails, and automatic disabling of access once the available privacy budget is exhausted \cite{cormode2018privacy}. Privacy budget tracking is increasingly required in regulated settings (e.g., hospital data governance platforms), reflecting that DP is not only a training-time guarantee but a limited resource that must be monitored throughout deployment \cite{66kaissis2021end}.

The amount of noise required to achieve DP depends on the sensitivity of the function $f$ being computed, which quantifies how much $f$ can change when a single input is modified.
For vector-valued outputs, the $\ell_2$-sensitivity of $f$ is:
\begin{equation}\label{l2sensitivity}
\Delta_2 f = \max_{\text{neighboring } D, D'} \| f(D) - f(D') \|_2
\end{equation}
This sensitivity bounds the required noise scale in mechanisms such as the Gaussian and Laplace mechanisms.

For real-valued outputs, the Gaussian mechanism adds noise $\eta \sim \mathcal{N}(0, \sigma^2)$ to $f(D)$:
\begin{equation}\label{eq:gaussian-mechanism}
\text{Release: } f(D) + \eta
\end{equation}
To achieve $(\varepsilon, \delta)$-DP, the standard deviation $\sigma$ must satisfy:
\begin{equation}
\sigma \geq \frac{\Delta_2 f \cdot \sqrt{2 \ln(1.25/\delta)}}{\epsilon}
\end{equation}

For $\epsilon$-DP, the Laplace mechanism adds noise from $\mathrm{Lap}(0, b)$, where:
\begin{equation}\label{SetScaleParameterInLaplace}
b = \frac{\Delta_1 f}{\epsilon}
\end{equation}
Here, $\Delta_1 f$ is the $\ell_1$-sensitivity of $f$.

When the output is non-numeric (e.g., a selected category), the exponential mechanism selects an output $r$ from a range $\mathcal{R}$ using a utility function $u(D, r)$:
\begin{equation}\label{sensitivityinexponentional}
\Pr[M(D) = r] \propto \exp\left(\frac{\epsilon \cdot u(D, r)}{2 \Delta u}\right)
\end{equation}
with sensitivity:
\begin{equation}
\Delta u = \max_{r \in \mathcal{R}} \max_{D, D'} | u(D, r) - u(D', r) |
\end{equation}
DP mechanisms can be applied at various stages of the learning pipeline, including the input~\cite{noise_at_input_1, noise_at_input_2, noise_at_input_3}, training process~\cite{DPSGD, noise_during_training_label}, and model output~\cite{noise_at_output_1, noise_at_output_2}. Among these, training-time DP has been shown to yield better accuracy and robustness~\cite{trade-offs}.

DL refers to models that learn hierarchical representations from data via neural networks with multiple layers~\cite{DeepLearningLeCun, DeepLearningSchmidhuber}. Training these models involves minimizing an empirical risk objective:
\begin{equation}
    J(\boldsymbol{\theta}) = \frac{1}{N} \sum_{i=1}^N L(\boldsymbol{\theta}; x_i, y_i)
\end{equation}
where $L$ is the loss function and $\boldsymbol{\theta}$ the model parameters. This is typically optimized using stochastic gradient descent (SGD).
Variants of SGD differ by the batch size used in each update. In practice, mini-batch SGD is most common:
\begin{equation}
    \boldsymbol{\theta}_{t+1} = \boldsymbol{\theta}_t - \eta \cdot \frac{1}{m} \sum_{j=1}^{m} \nabla_{\boldsymbol{\theta}} L(\boldsymbol{\theta}_t; x_{i_j}, y_{i_j})
\end{equation}
where $\eta$ is the learning rate and $m$ is the mini-batch size. This formulation enables efficient training with good generalization.

Deep neural networks are vulnerable to privacy attacks targeting the input data, gradients, or model parameters~\cite{ASurveyofPrivacyAttacksinMachineLearning}. Gradients are particularly sensitive, especially in federated settings where they are transmitted across devices~\cite{FederatedLearningSource}. Since gradients are computed directly from the input data, they can leak private information under adversarial scrutiny.
A common defense is to inject noise into gradients during training. DP-SGD~\cite{DPSGD} achieves this by modifying the gradient computation at each iteration. For each example $i$ in a mini-batch $\mathcal{B}$:
\begin{align}
\mathbf{g}_i &= \nabla_{\boldsymbol{\theta}} \mathcal{L}(\boldsymbol{\theta}; x_i, y_i) \label{eq:dpsgd1}\\
\tilde{\mathbf{g}}_i &= \mathbf{g}_i \cdot \min\left(1, \frac{C}{\|\mathbf{g}_i\|_2}\right) \label{eq:dpsgd2}\\
\bar{\mathbf{g}} &= \frac{1}{|\mathcal{B}|} \sum_{i \in \mathcal{B}} \tilde{\mathbf{g}}_i \label{eq:dpsgd3}\\
\hat{\mathbf{g}} &= \bar{\mathbf{g}} + \mathcal{N}(0, \sigma^2 C^2 \mathbf{I}) \label{eq:dpsgd4}\\
\boldsymbol{\theta} &\leftarrow \boldsymbol{\theta} - \eta \hat{\mathbf{g}} \label{eq:dpsgd5}
\end{align}
Here, $C$ is the clipping norm and $\sigma$ is the noise multiplier. Clipping limits any individual’s contribution, while the added noise ensures DP guarantees.


\subsection*{Overview of applications of DP in clinical DL}

\begin{figure}
    \centering
\includegraphics[width=\textwidth,height=\textheight,keepaspectratio]{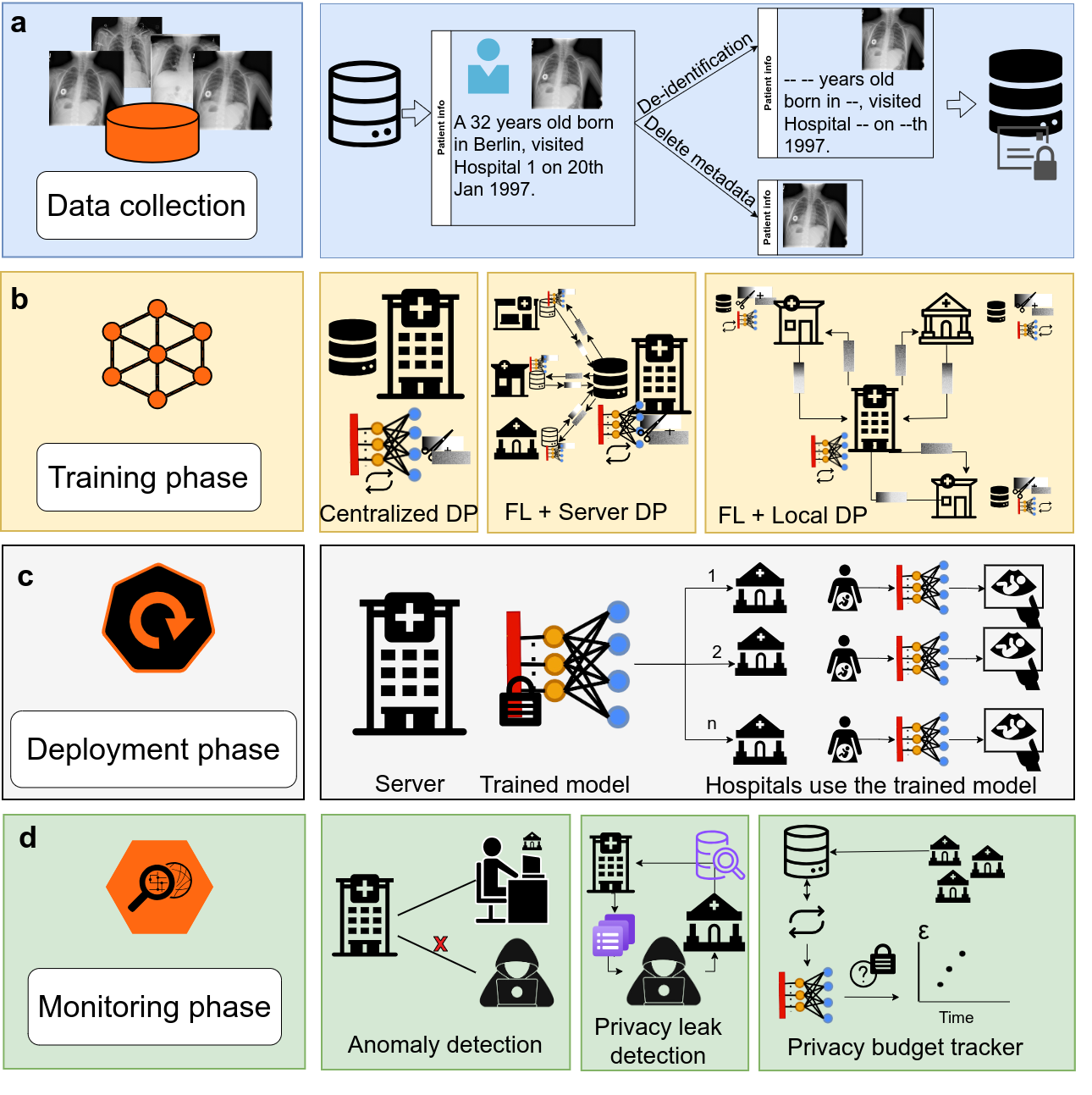}
\caption{Integrating differential privacy (DP) into the lifecycle of medical deep learning models. 
\textbf{(a)} Institutions curate data and remove direct identifiers or metadata before training begins. 
\textbf{(b)} DP can be applied in multiple configurations, including centralized DP, where all data are stored on a single server and DP-SGD is applied during training, federated learning (FL) with server-side DP, where institutions keep data locally, send model updates to a central server, and noise is added at the server before aggregation, and FL with local DP, where noise is added at each institution before sharing updates, providing stronger protection against an untrusted server.  
\textbf{(c)} The trained model is distributed to hospitals or clinical sites for inference.  
\textbf{(d)} After deployment, model behavior is monitored for privacy and security risks, including anomaly detection, privacy-leakage checks (e.g., membership inference signals), and tracking cumulative privacy budget when repeated queries or fine-tuning occur. Representative chest X-ray images are provided by the ChestX-ray14 dataset from NIH Clinical Center \cite{wang2017chestx}.}
\label{fig:deployment4}
\end{figure}

DP has been applied to a wide range of clinical use cases, spanning diverse data modalities, deployment settings, and downstream tasks. Table~\ref{tab:dpssgd_app} provides a summary of studies that applied DP via DP-SGD, the most widely adopted method in this domain \cite{DPSGD, de2022unlocking, chua2024private}. These studies encompass both centralized and FL settings, and cover modalities such as medical imaging (e.g., X-ray, CT, MRI), tabular data (e.g., EHR), biosignals (e.g., EEG, ECG), and multimodal inputs (Fig. \ref{fig:deployment4}).

\begin{table}[h]
    \centering
    \resizebox{\textwidth}{!}{%
        \begin{tabular}{p{3.2cm} >{\centering\arraybackslash}p{5.2cm} >{\centering\arraybackslash}p{1.9cm} >{\centering\arraybackslash}p{3.65cm} >{\centering\arraybackslash}p{2.0cm}}
            \toprule
            \textbf{Study} & \textbf{Modalities} & \textbf{Application} & \textbf{Downstream task}  & \textbf{Mech./account.} \\
            \midrule

            Adnan et al. \cite{36adnan2022federated} & Histopathology images & FL & Classification & GM/RDP \\
            
            Al Aziz et al. \cite{84DifferentiallyPrivateMedicalTextsGenerationUsingGenerativeNeuralNetworks} & EHR & Centralized & Tabular data generation & GM/MA \\
            
            Chilukoti et al. \cite{86chilukoti2025differentiallyprivatefinetunednfnet} & Histopathology images & Centralized & Classification & GM/RDP \\
            
            Chin-Cheong et al. \cite{74chin2020generation} & EHR & Centralized & Tabular data generation & GM/MA \\
            
            Daum et al. \cite{71OnDifferentiallyPrivate3DMedicalImageSynthesiswithControllableLatentDiffusionModels} & Cardiac cine MRI & Centralized & Image generation & GM/RDP \\
            
            DP-LPR \cite{25zhang2022secure} & Genomic data & Centralized & Feature selection & LM/NA  \\
            
            DP-SSLoRA \cite{24DP-SSLoRA} & Chest X-ray & Centralized & Classification & GM/RDP  \\
            
            DPFedSAM-Meas \cite{50DPFedSAM-Meas} & Chest X-ray & FL & Classification & GM, LM/RDP\\
            
            Fan et al.\cite{26fan2023mitigating} & Tabular data & Centralized & Regression & GM/RDP  \\
            
            FedGAN \cite{30nguyen2021federated} & Chest X-ray & FL & Classification & GM/NA  \\
            
            Guo et al. \cite{53Adifferentialprivacybasedprototypicalnetworkformedicaldatalearning} & Dermoscopic photography & Centralized & Classification & GM/NA \\
            
            Hatamizadeh et al. \cite{33hatamizadeh2023gradient} & Chest X-ray, fundoscopy & FL & Classification & GM/NA  \\
            
            Jones et al. \cite{21} & Time series & Centralized & Classification  & GM/RDP \\
            
            Kaess et al. \cite{88FairandPrivateCTContrastAgentDetection} & Chest CT & Centralized & Classification & GM/NA \\
            
            Khanna et al. \cite{32khanna2022privacy} & Genomic data & FL & Classification & GM/RDP  \\
            
            Lal et al. \cite{51DeepLearningClassificationofFetal} & Tabular data & Centralized & Classification & GM/MA \\
            
            Mehmood et al. \cite{47mehmood2024balancing} & Brain MRI & FL & Classification & GM/NA \\
            
            Mueller et al. \cite{46DifferentiallyPrivateGraphNeuralNetworksforWhole-GraphClassification} & Image-derived graph & Centralized & Classification & GM/RDP \\
            
            Odeyomi et al. \cite{48PreservingMedicalDatawithRenyiDifferentialPrivacy} & Chest X-ray & Centralized & Classification & GM/RDP \\
            
            P3SGD \cite{64wu2019p3sgd} & Histopathology images & Centralized & Classification & GM/MA \\
            
            Pan et al. \cite{63FedDP} & Histopathology images & FL & Segmentation & GM/NA \\
            
            PP-LDG \cite{59PP-LDG} & Abdominal CT & Centralized & Segmentation  & GM/MA \\
            
            Shiri et al. \cite{3DPFRCovid19} & Chest CT & FL & Regression  & GM/NA  \\
            
            Tayebi Arasteh et al. \cite{13TayebiDomainTransfer} & Chest X-ray & Centralized & Classification  & GM/RDP \\
            
            Tayebi Arasteh et al. \cite{22tayebi2024preserving} & Chest X-ray, abdominal CT & Centralized & Classification & GM/RDP  \\
            
            Tayebi Arasteh et al. \cite{72arasteh2024differentialprivacyenablesfair} & Pathological speech & Centralized & Classification & GM/RDP \\
            
            Torfi et al. \cite{7RDPCGAN} & EEG, EHR, ECG, biopsy & Centralized & Tabular data generation & GM/RDP \\
            
            Wang et al. \cite{28wang2022privacy} & Tabular data & FL & Classification & GM/MA  \\
            
            Ziller et al. \cite{35ziller2021medical} & Chest X-ray, abdominal CT, fundoscopy & Centralized & Classification, segmentation & GM/GDP  \\
            \bottomrule
        \end{tabular}%
    }
    \caption{Summary of studies applying DP-SGD in medical deep learning. Each entry includes the data modality, learning setting (centralized or FL), downstream task, and the DP mechanism and accounting method used. Only studies using DP-SGD are included. AD: Alzheimer's disease; ECG: electrocardiography; EEG: electroencephalography; EHR: electronic health record; FL: federated learning; GM: Gaussian mechanism; LM: Laplace mechanism; MA: moments accountant; NA: information not available; RDP: Rényi differential privacy.}
    \label{tab:dpssgd_app}
\end{table}

The reviewed studies apply DP-SGD across a broad range of medical data types, with imaging emerging as the most common modality. Chest X-rays \cite{wang2017chestx} dominate classification tasks \cite{13TayebiDomainTransfer, 22tayebi2024preserving, 24DP-SSLoRA, 30nguyen2021federated, 35ziller2021medical, 50DPFedSAM-Meas, 59PP-LDG, 88FairandPrivateCTContrastAgentDetection}, while MRI, particularly brain and cardiac cine MRI, is used in both classification and generative modeling applications \cite{47mehmood2024balancing, 71OnDifferentiallyPrivate3DMedicalImageSynthesiswithControllableLatentDiffusionModels}. Histopathology appears in several studies focusing on cancer classification and segmentation \cite{36adnan2022federated, 63FedDP, 64wu2019p3sgd, 86chilukoti2025differentiallyprivatefinetunednfnet}.
Beyond imaging, DP-SGD has been applied to structured data such as EHRs and tabular datasets \cite{26fan2023mitigating, 51DeepLearningClassificationofFetal, 74chin2020generation, 84DifferentiallyPrivateMedicalTextsGenerationUsingGenerativeNeuralNetworks}, genomic sequences \cite{25zhang2022secure, 32khanna2022privacy}, biosignals including EEG and ECG \cite{7RDPCGAN}, and pathological speech \cite{72arasteh2024differentialprivacyenablesfair}. Time series data is also explored in classification contexts \cite{21}, and some generative frameworks integrate multimodal data such as EEG, biopsy, and EHR \cite{7RDPCGAN}.

Classification remains the predominant downstream task \cite{13TayebiDomainTransfer, 22tayebi2024preserving, 24DP-SSLoRA, 30nguyen2021federated, 35ziller2021medical, 50DPFedSAM-Meas, 88FairandPrivateCTContrastAgentDetection, 47mehmood2024balancing, 86chilukoti2025differentiallyprivatefinetunednfnet, 26fan2023mitigating, 51DeepLearningClassificationofFetal, 32khanna2022privacy, 53Adifferentialprivacybasedprototypicalnetworkformedicaldatalearning, 21}, followed by generative modeling \cite{71OnDifferentiallyPrivate3DMedicalImageSynthesiswithControllableLatentDiffusionModels, 74chin2020generation, 84DifferentiallyPrivateMedicalTextsGenerationUsingGenerativeNeuralNetworks, 7RDPCGAN}. Despite growing interest, applications involving longitudinal data or multimodal fusion remain underexplored.

DP-SGD is employed in both centralized \cite{13TayebiDomainTransfer, 22tayebi2024preserving, 24DP-SSLoRA, 25zhang2022secure, 26fan2023mitigating, 35ziller2021medical, 46DifferentiallyPrivateGraphNeuralNetworksforWhole-GraphClassification, 48PreservingMedicalDatawithRenyiDifferentialPrivacy, 51DeepLearningClassificationofFetal, 53Adifferentialprivacybasedprototypicalnetworkformedicaldatalearning, 59PP-LDG, 64wu2019p3sgd, 71OnDifferentiallyPrivate3DMedicalImageSynthesiswithControllableLatentDiffusionModels, 72arasteh2024differentialprivacyenablesfair, 74chin2020generation, 84DifferentiallyPrivateMedicalTextsGenerationUsingGenerativeNeuralNetworks, 86chilukoti2025differentiallyprivatefinetunednfnet, 88FairandPrivateCTContrastAgentDetection, 7RDPCGAN, 21} and federated \cite{3DPFRCovid19, 28wang2022privacy, 30nguyen2021federated, 32khanna2022privacy, 33hatamizadeh2023gradient, 36adnan2022federated, 47mehmood2024balancing, 50DPFedSAM-Meas, 63FedDP} training pipelines. In centralized settings, models are trained on institutional data and privatized using DP-SGD, protecting against post-deployment leakage. These include imaging \cite{13TayebiDomainTransfer, 22tayebi2024preserving, 24DP-SSLoRA, 35ziller2021medical}, structured/tabular data \cite{26fan2023mitigating, 51DeepLearningClassificationofFetal}, and EHRs \cite{74chin2020generation, 84DifferentiallyPrivateMedicalTextsGenerationUsingGenerativeNeuralNetworks}.

FL-based applications involve multi-institutional collaboration, where local clients update models on-site and share only privatized gradients. These setups target imaging \cite{3DPFRCovid19, 30nguyen2021federated, 33hatamizadeh2023gradient}, genomics \cite{32khanna2022privacy}, and histopathology \cite{36adnan2022federated, 63FedDP}. Despite their promise, FL studies often omit key DP parameters, complicating reproducibility.
While FL offers inherent privacy through data decentralization \cite{tayebi2023collaborative, tayebi2023enhancing, FederatedLearningSource}, centralized DP-SGD remains vital for models intended for public sharing or external evaluation. The versatility of DP-SGD across both paradigms underscores its relevance to privacy-preserving clinical AI.


The Gaussian mechanism (GM), paired with RDP, is the dominant configuration in DP-SGD implementations, offering favorable composition properties and practical support through libraries like Opacus \cite{opacus_neurips2021} and TensorFlow Privacy \cite{dahl2018private, papernot2019machine}. Most reviewed works adopt GM with RDP for per-sample gradient perturbation.

Nonetheless, reporting practices are inconsistent. Some studies provide complete details, including $(\epsilon, \delta)$ values, clipping norms, and noise multipliers \cite{13TayebiDomainTransfer, 22tayebi2024preserving, 26fan2023mitigating, 46DifferentiallyPrivateGraphNeuralNetworksforWhole-GraphClassification, 72arasteh2024differentialprivacyenablesfair}, while others omit the accounting mechanism or final budget \cite{3DPFRCovid19, 30nguyen2021federated, 33hatamizadeh2023gradient}. A few exceptions use the Laplace mechanism (LM) \cite{25zhang2022secure} or combine GM and LM \cite{50DPFedSAM-Meas}. The moments accountant (MA) is used in some centralized and FL settings \cite{28wang2022privacy, 51DeepLearningClassificationofFetal}.
Privacy budgets range from conservative ($\epsilon \approx 1$) to more relaxed ($\epsilon \approx 10$). However, few studies systematically explore multiple $\epsilon$ levels, limiting our understanding of utility degradation under stronger privacy guarantees.
Overall, while GM and RDP dominate, standardization in reporting mechanisms, parameters, and accounting tools remains a critical need for enabling rigorous comparisons and reproducibility across studies.

\subsection*{Design choices for privacy-utility tradeoffs}

Introducing DP into DL workflows inevitably introduces tradeoffs between model utility and privacy guarantees. The severity of this tradeoff is highly sensitive to several key design decisions, including architecture choice, normalization strategy, and data augmentation. Table~\ref{tab:utility_design} summarizes these technical configurations across reviewed studies.

\begin{table}[h]
    \centering
    \resizebox{\textwidth}{!}{%
        \begin{tabular}{p{3.3cm} >{\centering\arraybackslash}p{3.6cm} >{\centering\arraybackslash}p{3.75cm} >{\centering\arraybackslash}p{1.9cm} >{\centering\arraybackslash}p{1.5cm}}
            \toprule
            \textbf{Study} & \textbf{Modalities} & \textbf{Architectures} & \textbf{Normalization} & \textbf{Data aug.} \\
            \midrule
            
            Adnan et al. \cite{36adnan2022federated} & Histopathology images & MEM-MIL & NA & NA \\

            Beaulieu-Jones et al. \cite{21} & Time series & AC-GAN \cite{odena2017conditional} &  NA & NA  \\

            Chin-Cheong et al. \cite{74chin2020generation} & EHR & Wasserstein GAN & MSN & NA \\

            Chilukoti et al. \cite{86chilukoti2025differentiallyprivatefinetunednfnet} & Histopathology & NF-Net \cite{brock2021high} & NA & NA \\

            Daum et al. \cite{71OnDifferentiallyPrivate3DMedicalImageSynthesiswithControllableLatentDiffusionModels} & Cardiac cine MRI & Latent diffusion models \cite{rombach2022high} & NA & NA \\

            Fan et al. \cite{26fan2023mitigating} & Tabular data &  MLP & NA & NA \\

            Guo, Yu et al. \cite{53Adifferentialprivacybasedprototypicalnetworkformedicaldatalearning} & Dermoscopic photography & NA & GN & NA \\

            Kaess et al. \cite{88FairandPrivateCTContrastAgentDetection} & Chest CT & ResNet-9 \cite{he2016deep} & Scale norm & NA \\

            Khanna et al. \cite{32khanna2022privacy} & Genomic data & MLP & NA & NA \\

            Mehmood et al. \cite{47mehmood2024balancing} & Brain MRI & InceptionV3 \cite{szegedy2015going} & NA & Affine \\

            Mueller et al. \cite{46DifferentiallyPrivateGraphNeuralNetworksforWhole-GraphClassification} & Image-derived graph & Graph neural networks& Instance norm &  NA\\

            Nguyen et al.\cite{30nguyen2021federated} &  Chest X-ray & GAN & NA & NA \\

            Riess et al. \cite{73Complex-ValuedFederatedLearningwithDifferentialPrivacy} &  Brain MRI & NA & GN & NA \\

            Tayebi Arasteh et al. \cite{13TayebiDomainTransfer} & Chest X-ray & ResNet-9 & GN & None \\

            Tayebi Arasteh et al. \cite{22tayebi2024preserving} & Chest X-ray, abdominal CT & ResNet-9 & GN & None \\

            Tayebi Arasteh et al. \cite{72arasteh2024differentialprivacyenablesfair} & Pathological speech & ResNet-18 & GN & None \\

            Yan et al. \cite{24DP-SSLoRA} & Chest X-ray & ResNet-18 & None & NA \\

            Zhang et al. \cite{25zhang2022secure} & Genomic data & Custom CNN & NA & NA \\

            Ziller et al. \cite{35ziller2021medical} & Chest X-ray, abdominal CT & VGG-11, custom U-Net & None & Affine \\
            
            \bottomrule
        \end{tabular}%
    }
    \caption{Model design strategies for managing privacy-utility tradeoffs in DP-SGD applications. The table lists architectures, data normalization techniques, and data augmentation practices used in studies applying DP-SGD to medical deep learning. GN: group normalization; MEM-MIL: multi-expert MIL; MLP: multilayer perceptron;  MSN: mode-specific normalization; NA: information not available.}
    \label{tab:utility_design}
\end{table}

Architecture selection significantly influences a model’s resilience to the noise and clipping operations imposed by DP-SGD. Compact CNNs such as ResNet-9 and ResNet-18 \cite{he2016deep} were the most widely adopted across classification tasks in chest X-ray, CT, and speech data \cite{13TayebiDomainTransfer, 22tayebi2024preserving, 24DP-SSLoRA, 72arasteh2024differentialprivacyenablesfair, 88FairandPrivateCTContrastAgentDetection}. Other lightweight backbones like VGG-11 and custom U-Net variants were applied in medical imaging tasks \cite{35ziller2021medical}, while NF-Net was used in histopathology \cite{86chilukoti2025differentiallyprivatefinetunednfnet}, and InceptionV3 in brain MRI \cite{47mehmood2024balancing}.

Generative models used GAN architectures, including Wasserstein GANs for EHR \cite{74chin2020generation}, AC-GANs for time series \cite{21}, and standard GANs in FL imaging tasks \cite{30nguyen2021federated}. More recent approaches explored latent diffusion models for cardiac MRI synthesis \cite{71OnDifferentiallyPrivate3DMedicalImageSynthesiswithControllableLatentDiffusionModels}. MLPs remain prevalent for tabular and genomic data \cite{26fan2023mitigating, 32khanna2022privacy}, while specialized networks such as graph neural networks (GNNs) \cite{46DifferentiallyPrivateGraphNeuralNetworksforWhole-GraphClassification} and MEM-MIL \cite{36adnan2022federated} were applied in niche contexts like graph classification and histopathology.
Notably, vision transformers (ViTs) \cite{mainvit} and other large-scale models were absent from this set, echoing findings in separate comparisons that such architectures are more sensitive to DP noise and tend to underperform under strong privacy constraints.

Batch normalization (BN) \cite{ioffe2015batch}, which depends on mini-batch statistics, is generally avoided in DP-SGD because it interferes with per-sample gradient computation \cite{22tayebi2024preserving, kaissis2020secure, 13TayebiDomainTransfer, DPSGD, 72arasteh2024differentialprivacyenablesfair}. Among studies reporting normalization explicitly, group normalization (GN) \cite{wu2018group} was most common, used in both classification and speech tasks \cite{13TayebiDomainTransfer, 22tayebi2024preserving, 72arasteh2024differentialprivacyenablesfair, 53Adifferentialprivacybasedprototypicalnetworkformedicaldatalearning, 73Complex-ValuedFederatedLearningwithDifferentialPrivacy}. GN avoids batch dependencies and offers more stable training under DP noise.
Other strategies include instance normalization for graph neural networks \cite{46DifferentiallyPrivateGraphNeuralNetworksforWhole-GraphClassification}, mode-specific normalization (MSN) \cite{deecke2018mode} for generative EHR models \cite{74chin2020generation}, and scale normalization in CT classification \cite{88FairandPrivateCTContrastAgentDetection}. Several studies, particularly those using GANs, MLPs, or CNNs in tabular or genomic settings, did not specify any normalization approach \cite{21, 25zhang2022secure, 26fan2023mitigating, 30nguyen2021federated, 32khanna2022privacy, 36adnan2022federated}. Others reported disabling normalization entirely \cite{24DP-SSLoRA, 35ziller2021medical}, likely to avoid incompatibility with gradient clipping.
Overall, there is no universal best practice, though GN appears robust across diverse tasks. Systematic benchmarking of normalization techniques under DP constraints remains a key direction for future research.


Data augmentation, a staple of conventional DL, is rarely applied under DP-SGD. Among reviewed studies, only two used basic affine transformations, one for brain MRI classification \cite{47mehmood2024balancing} and one for multi-modal segmentation \cite{35ziller2021medical}. Most others either explicitly avoided augmentation \cite{13TayebiDomainTransfer, 22tayebi2024preserving, 72arasteh2024differentialprivacyenablesfair} or did not report it.
This reluctance stems from the risk of increasing gradient variance, which exacerbates training instability under noisy DP updates. Augmentations can alter input distributions in ways that interfere with the gradient signal, especially in small medical datasets with high class imbalance \cite{22tayebi2024preserving}.
The lack of systematic evaluation limits conclusions about which augmentations, if any, are compatible with DP-SGD. There is a clear need for controlled experiments testing standard and domain-specific augmentation strategies under DP conditions.


Pretraining is a consistent and effective strategy for mitigating performance loss under DP-SGD \cite{13TayebiDomainTransfer, 22tayebi2024preserving, 72arasteh2024differentialprivacyenablesfair}. By starting from pretrained weights, models avoid the fragile convergence dynamics associated with noisy training from scratch.
Tayebi Arasteh et al.\ \cite{22tayebi2024preserving} demonstrated that DP-SGD models trained from scratch failed to learn meaningful features, whereas pretrained ResNet-9 models retained high accuracy. In a related study, pretrained models on chest X-ray and CT were shown to maintain near non-private performance under tight $\epsilon$ budgets \cite{22tayebi2024preserving}. Pretraining on general-domain speech corpora also improved DP training on pathological speech classification \cite{72arasteh2024differentialprivacyenablesfair}, despite cross-domain mismatch. Similarly, ImageNet-initialized models improved performance in brain MRI classification under DP \cite{47mehmood2024balancing}.
Pretraining is not only helpful but often necessary for training deeper or more expressive models under DP-SGD \cite{72arasteh2024differentialprivacyenablesfair}. Explicit reporting and standardization of pretraining strategies are vital for reproducibility and benchmarking.


A core challenge in differentially private DL is navigating the tradeoff between model utility and privacy \cite{22tayebi2024preserving, kaissis2020secure}. As summarized in Table~\ref{tab:utility_results}, studies vary widely in how they report performance under different privacy budgets, with only a subset evaluating multiple $\epsilon$ levels. This inconsistency limits comparative understanding across clinical modalities and tasks.
At very low privacy budgets ($\epsilon \approx 1$), the magnitude of injected noise often leads to significant performance degradation. For example, Tayebi Arasteh et al.\ \cite{22tayebi2024preserving} observed an AUC drop from 89.7\% to 84.0\% on chest X-rays and from 99.7\% to 92.0\% on abdominal CT when applying $\epsilon = 0.5$. In pathological speech classification, accuracy fell from 99.1\% to 88.3\% at $\epsilon = 0.9$ \cite{72arasteh2024differentialprivacyenablesfair}, and in dermoscopy, accuracy declined from 69.1\% to ~43\% at $\epsilon = 3$ \cite{53Adifferentialprivacybasedprototypicalnetworkformedicaldatalearning}. Similar trends were reported for generative tasks, such as cardiac MRI synthesis, where Fréchet inception distance (FID) worsened substantially under $\epsilon = 1$ \cite{71OnDifferentiallyPrivate3DMedicalImageSynthesiswithControllableLatentDiffusionModels}.
However, there are exceptions. In a multi-institutional study on domain transfer, Tayebi Arasteh et al.\ \cite{13TayebiDomainTransfer} showed that external validation performance remained nearly unchanged across $\epsilon \approx 1$–10. This suggests that DP noise may act as a regularizer, benefiting generalization even as in-domain performance drops.

Most reviewed studies operate in the strong privacy regime ($\epsilon \approx 5$–10), where performance generally remains close to non-private baselines. For instance, chest X-ray AUC dropped modestly from 89.7\% to 87.4\% at $\epsilon = 7.9$, and CT classification remained above 99\% even at $\epsilon = 8$ \cite{22tayebi2024preserving}. In MRI \cite{47mehmood2024balancing} and genomic data \cite{32khanna2022privacy}, performance losses at $\epsilon = 10$ were similarly modest.
Nonetheless, more fragile domains show greater vulnerability to DP noise. Studies in histopathology \cite{36adnan2022federated}, dermoscopy \cite{53Adifferentialprivacybasedprototypicalnetworkformedicaldatalearning}, and speech \cite{72arasteh2024differentialprivacyenablesfair} report larger performance drops, reflecting higher input variance, smaller datasets, and more complex learning tasks. Tabular and sensor-based applications, e.g., in mental health or activity monitoring, also showed substantial utility loss at lower $\epsilon$ values \cite{31AZIZ2022104113, 60DifferentialPrivateFederatedTransferLearningforMentalHealthMonitoringinEverydaySettings}.
These trends emphasize the importance of modality-specific tuning \cite{72arasteh2024differentialprivacyenablesfair}: modalities like chest imaging are more robust under DP-SGD \cite{13TayebiDomainTransfer, 22tayebi2024preserving}, while others require additional regularization, pretraining, or architectural adaptation to maintain clinical viability.

While most studies report $\epsilon$, the value of $\delta$, which bounds the probability of privacy failure, is often missing or inconsistently stated. Best practice recommends $\delta \leq 1/n$, where $n$ is the dataset size. Many studies follow this, setting $\delta$ between $10^{-5}$ and $10^{-6}$, but others omit it or use heuristic values without justification.

Reporting gaps extend beyond privacy parameters. Performance under multiple $\epsilon$ levels is rarely benchmarked, and core implementation details, such as clipping norm, noise multiplier, accounting method (e.g., RDP or moments accountant~\cite{DPSGD}), and number of training epochs, are frequently missing. This limits reproducibility and cross-study comparison.
This issue is particularly acute in FL settings, where DP is often applied to distributed model updates. Although FL introduces additional privacy through decentralization \cite{tayebi2023collaborative, tayebi2023enhancing, FederatedLearningSource, tayebiarasteh23_interspeech}, few studies evaluate how DP noise affects utility across varying budgets in such settings, hindering generalizability.

To address these gaps, we recommend several reporting standards for future work. (i) Researchers should report both non-private and private model performance to provide clear context for evaluating the impact of privacy mechanisms. (ii) The values of key privacy parameters, including $\epsilon$, $\delta$, and the accounting method used (e.g., RDP or moments accountant), should be explicitly stated to ensure interpretability and comparability. (iii) Performance metrics should be reported across multiple $\epsilon$ levels to characterize the tradeoff between privacy and utility. (iv) Finally, studies should document core model and training details, including architecture type, dataset size, pretraining strategies, and other relevant hyperparameters. Adopting these practices will improve transparency, foster reproducibility, and support more rigorous benchmarking of privacy-preserving models for clinical deployment.

\begin{table}[h]
    \centering
    \resizebox{\textwidth}{!}{%
        \begin{tabular}{p{3.3cm} >{\centering\arraybackslash}p{3.4cm} >{\centering\arraybackslash}p{2.8cm} >{\centering\arraybackslash}p{3.5cm} >{\centering\arraybackslash}p{3.5cm}}
            \toprule
            \textbf{Study} & \textbf{Modality} & \textbf{Non-private} & \textbf{Very low budget} & \textbf{Low budget} \\
            \midrule
        
            Adnan et al. \cite{36adnan2022federated} & Histopathology images &  Accuracy = $82\% $ & $\epsilon = 2.9$/Accuracy = $81\%$ & $\epsilon = 10$/Accuracy = $78\%$ \\

            Al Aziz et al. \cite{31AZIZ2022104113} & Tabular data &  AUC = \textasciitilde$ 98\%$ & $\epsilon = 1/$AUC = $70\%$ & $\epsilon = 10/$AUC = $92\%/$ \\

            Alsenani et al. \cite{23FAItH} & Sensor data & Accuracy = $30\%$ &  NA &  $\epsilon = 10$/Accuracy = $27\%$ \\

            Beaulieu-Jones et al. \cite{21} & Time series  & Spearman = $0.96$ &  $\epsilon=3$/Spearman = $0.91$ & NA \\

            Chin-Cheong et al. \cite{74chin2020generation} & EHR &  AUC = $80.0\% $ & $\epsilon = 1$/AUC = $66.1\% $ & $\epsilon = 10$/AUC = $67.8\%$ \\

            Daum et al. \cite{71OnDifferentiallyPrivate3DMedicalImageSynthesiswithControllableLatentDiffusionModels} & Cardiac cine MRI &  FID = $15.4 $ & $\epsilon = 1$/FID = $29.8 $ & $\epsilon = 10$/FID = $26.8 $ \\

            Fan et al. \cite{26fan2023mitigating} & Tabular data & IBS = \textasciitilde$17$ & $\epsilon = 1$/IBS = \textasciitilde$21 $ & $\epsilon = 8$/IBS = \textasciitilde$18 $  \\

            Fu et al. \cite{82DP-MLD} & EEG, sensor data & Accuracy = $99.3\% $ & $\epsilon = 1$/Accuracy = $98.7\%$ & NA\\

            Gong et al. \cite{27gong2023federated} & Time series & Accuracy = $88.5\%$ & $\epsilon = 2$/Accuracy = $75.3\% $ & $\epsilon = 10$/Accuracy = $78.2\% $  \\

            Guo, Yu et al. \cite{53Adifferentialprivacybasedprototypicalnetworkformedicaldatalearning} & Dermoscopic photography &  Accuracy = $69.1\%$ & $\epsilon = 3$/Accuracy = \textasciitilde$43\% $ & $\epsilon = 10$/Accuracy = $61.2\% $ \\

            Kaess et al. \cite{88FairandPrivateCTContrastAgentDetection} & Chest CT &  Accuracy = $99.5\%$ & NA & $\epsilon =  8$ /Accuracy = $97.4\%$\\

            Mehmood et al. \cite{47mehmood2024balancing} & Brain MRI &Accuracy = $88\%$& NA & $\epsilon = 5.6$/Accuracy = $78\%$\\

            Ming Y et al. \cite{42GWON2024107738} & Tabular EMR & DWS =$0.0041 $ & $\epsilon = 1$/DWS = $0.05$ & $\epsilon =10$/DWS = $0.0061$ \\

            Mueller et al. \cite{46DifferentiallyPrivateGraphNeuralNetworksforWhole-GraphClassification} & Image-derived graph & AUC = $98.5\%$ & $\epsilon = 1$/AUC = $95.1\% $ & $\epsilon = 10$/AUC = $97.2\% $\\

            Nguyen et al. \cite{30nguyen2021federated} &  Chest X-ray & Accuracy = $99\% $ & $\epsilon = 0.3$/Accuracy = \textasciitilde$90\% $ & NA  \\

            Odeyomi et al. \cite{49yuan2019collaborative} & Chest X-ray  & Accuracy = $92\% $ & $\epsilon = 2$/Accuracy = $88\% $ & $\epsilon = 8$/Accuracy = $90\% $\\

            Riess et al. \cite{73Complex-ValuedFederatedLearningwithDifferentialPrivacy} & Brain MRI &   Accuracy = $90.1\% $ & $\epsilon = 1$/Accuracy = $81.6\% $ & $\epsilon = 10$/Accuracy = $90\% $ \\

            Sun et al. \cite{39SUN2023104404} & Tabular data & DOP = \textasciitilde$0.75 $ & $\epsilon = 1$/DOP = \textasciitilde$1.2 $ & $\epsilon = 10$/DOP = \textasciitilde$1.2 $\\

            Tang et al. \cite{34tang2024personalized} & Tabular data  &  F1-score = $19.6\%$ & $\epsilon = 1$/F1-score = $19.79\%$ & NA \\

            Tayebi Arasteh et al. \cite{13TayebiDomainTransfer} & Chest X-ray & AUC$=68-90\%$ & $\epsilon \approx 1$/AUC$=67-89\%$ & $\epsilon \approx 10$/AUC$ = 67-89\%$ \\

            Tayebi Arasteh et al. \cite{22tayebi2024preserving} & Chest X-ray & AUC$=89.7\%$ & $\epsilon =0.5$/AUC$=84.0\%$ & $\epsilon =7.9$/AUC$=87.4\%$ \\

            Tayebi Arasteh et al. \cite{22tayebi2024preserving} & Abdominal CT & AUC$=99.7\%$ & $\epsilon =0.5$/AUC$=92\%$ & $\epsilon =8.0$/AUC$=99.3\%$ \\

            Tayebi Arasteh et al. \cite{72arasteh2024differentialprivacyenablesfair} & Pathological speech & Accuracy = $99.1\%$ & $\epsilon = 0.9$/Accuracy = $88.3\%$ & $\epsilon = 7.5$/Accuracy = $95.3\% $ \\

            Wang et al. \cite{60DifferentialPrivateFederatedTransferLearningforMentalHealthMonitoringinEverydaySettings} & Sensor data & AUC = $59\% $ & $\epsilon = 1$/AUC = $56\% $ & NA\\

            Yan et al. \cite{24DP-SSLoRA} & Chest X-ray & AUC = $99\%$ & $\epsilon = 2$/AUC = $96.5\% $ & $\epsilon = 10$/AUC = $98\% $\\

            Zhang et al. \cite{25zhang2022secure} & DNA-sequence  & Accuracy = $96\%$ & $\epsilon = 1$/Accuracy = \textasciitilde$87\% $ & NA  \\

            Zhang et al. \cite{62BalancingCentralizedandLocalDifferentialPrivacyinPneumoniaDiagnosis} & Chest X-ray &  Accuracy = $92.9\%$ & $\epsilon = 1$/Accuracy = $84.7\% $ & $\epsilon = 10$/Accuracy = $91.5\% $ \\

            Ziller et al. \cite{35ziller2021medical} & Chest X-ray &  AUC = $96\% $ & $\epsilon = 0.5$/AUC = $84\%$ & NA \\

            Ziller et al. \cite{35ziller2021medical} & Abdominal CT &  AUC = $96\% $ & $\epsilon = 0.5$/Dice= $94\%$ & NA \\
            
            \bottomrule
        \end{tabular}%
    }
    \caption{Reported privacy-utility tradeoffs across varying privacy budgets. This table summarizes model performance in both non-private settings and under different levels of protection, focusing on very strong protection (very low: $\epsilon \approx 1$) and strong protection (low: $\epsilon \approx 5$–10). Note that some studies applied DP to more than one modality or task; in such cases, each modality-task pair is listed as a separate row. Metrics vary by task and include accuracy, area under the receiver operating characteristic curve (AUC), F1-score, Spearman correlation, integrated Brier score (IBS), distance of optimal calibration (DOP), distance between Wasserstein barycenters (DWS), Dice score, and Fréchet inception distance (FID), as reported by each study. NA: Information not available.}
    \label{tab:utility_results}
\end{table}

\subsection*{Fairness considerations in DP-based medical DL}

While DP is designed to protect individual data, its application in DL can unintentionally exacerbate performance disparities across demographic subgroups \cite{tran2021decision, 22tayebi2024preserving, cummings2019compatibility}. This is particularly concerning in healthcare, where systemic biases already exist, and privacy mechanisms that disproportionately impact underrepresented groups can further entrench inequities. Table~\ref{tab:fairness} summarizes reviewed studies that assess the fairness implications of DP in medical DL, the subgroup attributes evaluated, and the metrics used.

\begin{table}[h]
    \centering
    \resizebox{\textwidth}{!}{%
        \begin{tabular}{p{4.2cm} >{\centering\arraybackslash}p{3.6cm} >{\centering\arraybackslash}p{6cm} >{\centering\arraybackslash}p{2.5cm}}
            \toprule
            \textbf{Study} & \textbf{Modalities} & \textbf{Use case} & \textbf{Fairness groups} \\
            \midrule

            \textbf{DP privacy-fairness} \\
            \midrule

            Chin-Cheong et al. \cite{74chin2020generation} & EHR & Text/tabular data generation & Age, sex \\
            
            Kaess et al. \cite{88FairandPrivateCTContrastAgentDetection} & Chest CT & Classification & Race, age, sex \\

            Tayebi Arasteh et al. \cite{13TayebiDomainTransfer} & Chest X-ray & Classification & Age, sex \\
            
            Tayebi Arasteh et al. \cite{22tayebi2024preserving} & Chest X-ray, abdominal CT & Classification & Age, sex \\
            
            Tayebi Arasteh et al. \cite{72arasteh2024differentialprivacyenablesfair} & Pathological speech & Classification & Age, sex \\

            \midrule
            \textbf{Fairness metrics} \\
            \midrule

            Average odds difference \cite{hardt2016equality} & Any & Mean gap in TPR and FPR across groups & Any \\

            Disparate impact ratio \cite{eeoc1978uniform, feldman2015certifying} & Any & Ratio of PPRs & Any \\

            Equal accuracy \cite{hardt2016equality} & Any & Difference in overall accuracy across groups & Any \\

            Equal opportunity \cite{hardt2016equality} & Any & Difference in TPR across groups & Any \\

            Equalized odds difference \cite{hardt2016equality} & Any & Difference in both TPR and FPR & Any \\

            FDR parity \cite{chouldechova2017fair} & Any & Difference in FDR across groups & Any \\

            FOR parity \cite{chouldechova2017fair} & Any & Difference in FOR across groups & Any \\

            Group-wise calibration \cite{pleiss2017fairness} & Any & Prediction-outcome gap across groups & Any \\

            Overdiagnosis rate \cite{obermeyer2019dissecting} & Image, tabular & Excess positive predictions across groups & Age, sex \\

            Predictive equality \cite{hardt2016equality} & Any & Difference in FPR across groups & Any \\

            PtD \cite{dwork2012fairness, 22tayebi2024preserving} & Any & Gap in PPRs across groups & Any \\

            Subgroup AUC \cite{22tayebi2024preserving} & Any, especially image & AUC reported per group to assess disparities & Age, sex, race \\

            Treatment equality \cite{corbett2017algorithmic} & Any & Ratio of FN to FP across groups & Any \\

            Underdiagnosis rate \cite{obermeyer2019dissecting} & Image, tabular & Difference in FNRs across groups & Age, sex \\

            \bottomrule
        \end{tabular}%
    }
    \caption{Summary of privacy-fairness studies and fairness evaluation metrics. The upper section lists studies that explicitly investigate fairness impacts of applying DP in medical deep learning, including use cases and demographic groups analyzed. The lower section summarizes fairness metrics reported across studies, covering various dimensions of group-level equity. "Any" in the modalities or fairness groups column indicates that the metric is applicable across any modality or protected attribute (e.g., age, sex, race). AUC: area under the receiver operating characteristic curve; EHR: electronic health record; FDR: false discovery rate; FNR: false negative rate; FOR: false omission rate; FPR: false positive rate; PPR: positive prediction rate; PtD: statistical parity difference; TPR: true positive rate.}
    \label{tab:fairness}
\end{table}

Only a limited number of studies explicitly examine fairness under DP constraints \cite{74chin2020generation, 88FairandPrivateCTContrastAgentDetection, 13TayebiDomainTransfer, 22tayebi2024preserving, 72arasteh2024differentialprivacyenablesfair}. These focus primarily on group-level disparities in classification tasks across attributes such as age, sex, and race. For instance, Tayebi Arasteh et al.\ evaluated fairness across sex and age in chest X-ray \cite{22tayebi2024preserving} and speech classification \cite{72arasteh2024differentialprivacyenablesfair}; Kaess et al.\ \cite{88FairandPrivateCTContrastAgentDetection} examined racial disparities in chest CT; and Chin-Cheong et al.\ \cite{74chin2020generation} analyzed synthetic EHR outputs.
Findings suggest that the impact of DP is highly context-dependent. In some cases, privacy noise interacts with data imbalance to disproportionately degrade performance for certain groups, especially minorities with fewer examples. For example, DP constraints in speech classification disproportionately harmed age groups \cite{72arasteh2024differentialprivacyenablesfair}.
Despite these insights, most studies report fairness at only a single privacy level. Only a few works, such as \cite{13TayebiDomainTransfer, 22tayebi2024preserving, 72arasteh2024differentialprivacyenablesfair}, systematically analyze fairness across multiple $\epsilon$ values, limiting our understanding of how fairness evolves as privacy constraints tighten.
Moreover, fairness is rarely a design objective. While certain configurations (e.g., ResNet backbones or normalization with GN) were found to reduce subgroup variance \cite{22tayebi2024preserving, 72arasteh2024differentialprivacyenablesfair}, these choices were primarily utility-driven. Fairness-aware techniques like constraint-based learning, adversarial debiasing, or subgroup reweighting are almost entirely absent.

Broader gaps remain. Most evaluations consider only binary attributes, overlooking intersectional identities (e.g., older female patients), and focus almost exclusively on classification, neglecting regression and survival tasks. Additionally, the lack of demographically diverse benchmarks hinders fairness auditing under DP constraints. Progress will require treating fairness as a first-class design goal in DP pipelines, and systematically reporting subgroup performance across varying privacy levels.


Fairness auditing is essential when applying DP to medical DL, as noise can distort performance differently across groups \cite{67ziller2024reconciling}. Reviewed studies primarily assess group-level metrics based on protected attributes such as age, sex, or race. Table~\ref{tab:fairness} summarizes the main metrics used.
Most fairness evaluations focus on group-level comparisons, typically assessing disparities across sex, age, or race. The most frequently reported metric is statistical parity difference (PtD) \cite{dwork2012fairness, 22tayebi2024preserving, 13TayebiDomainTransfer, 72arasteh2024differentialprivacyenablesfair, tayebi2024addressing, 87xing2025achieving}, which measures the difference in the proportion of positive predictions between two subgroups:

\begin{equation}
\text{PtD} = \Pr(\hat{Y} = 1 \mid A = a) - \Pr(\hat{Y} = 1 \mid A = b)
\label{eq:ptd}
\end{equation}
Here, $\hat{Y}$ is the model prediction, $A$ is a protected attribute (e.g., sex), and $a$, $b$ are two group values (e.g., male and female).
A related metric is the disparate impact ratio (DIR) \cite{eeoc1978uniform, feldman2015certifying}, which compares the rates of favorable outcomes as a ratio:

\begin{equation}
\text{DIR} = \frac{\Pr(\hat{Y} = 1 \mid A = a)}{\Pr(\hat{Y} = 1 \mid A = b)}
\label{eq:dir}
\end{equation}

Another widely used metric is the equal opportunity difference \cite{hardt2016equality, 72arasteh2024differentialprivacyenablesfair}, which focuses on differences in true positive rates (TPRs) between groups:

\begin{equation}
\Delta \text{EqOp} = \Pr(\hat{Y} = 1 \mid Y = 1, A = a) - \Pr(\hat{Y} = 1 \mid Y = 1, A = b)
\label{eq:eqop}
\end{equation}
In this case, $Y$ is the ground truth label, with $Y = 1$ indicating a positive case. Equalized odds \cite{hardt2016equality} extends this concept by considering both true positive and false positive rates:

\begin{align}
\Delta \text{EOD} =\ & \left| \Pr(\hat{Y} = 1 \mid Y = 1, A = a) - \Pr(\hat{Y} = 1 \mid Y = 1, A = b) \right| \notag \\
&+ \left| \Pr(\hat{Y} = 1 \mid Y = 0, A = a) - \Pr(\hat{Y} = 1 \mid Y = 0, A = b) \right|
\label{eq:eqodds}
\end{align}
Predictive equality \cite{hardt2016equality}, by contrast, focuses only on disparities in false positive rates:

\begin{equation}
\Delta \text{FPR} = \Pr(\hat{Y} = 1 \mid Y = 0, A = a) - \Pr(\hat{Y} = 1 \mid Y = 0, A = b)
\label{eq:pred_eq}
\end{equation}
Treatment equality \cite{corbett2017algorithmic} compares the balance between false negatives (FN) and false positives (FP) across groups:

\begin{equation}
\text{Treatment Equality} = \frac{\text{FN}_{a} / \text{FP}_{a}}{\text{FN}_{b} / \text{FP}_{b}}
\label{eq:treatment_eq}
\end{equation}

In addition to general-purpose fairness metrics, several studies employed domain-specific measures such as underdiagnosis rate \cite{obermeyer2019dissecting} (differences in false negative rates across groups), overdiagnosis rate (difference in false positive rate across groups), and subgroup AUC \cite{22tayebi2024preserving, 72arasteh2024differentialprivacyenablesfair, 13TayebiDomainTransfer}. While these do not require formal equations, they are particularly important in clinical settings, where different types of errors carry different consequences for patient care. 
Other metrics such as equal accuracy \cite{hardt2016equality}, group-wise calibration \cite{pleiss2017fairness}, false discovery rate parity and false omission rate parity \cite{chouldechova2017fair} were also used in a small subset of studies, though less frequently than PtD and equal opportunity. 
While these metrics are useful, they must be interpreted in context. For example, parity in outcomes across age groups may be inappropriate when prevalence rates differ biologically. Furthermore, DP-induced noise may distort probability calibration, reducing the reliability of threshold-based or probabilistic fairness measures.

To ensure robustness, future studies should adopt a standardized approach to fairness auditing under DP: (i) report multiple fairness metrics to capture different dimensions of disparity; (ii) justify the choice of protected attributes based on clinical relevance; (iii) disaggregate results by subgroup across multiple $\epsilon$ values; and (iv) apply calibration or post hoc adjustments when using probability-based metrics. The development of benchmark datasets with detailed demographic annotations is also critical to advance fair, privacy-preserving AI in medicine.

\subsection*{Alternative privacy mechanisms beyond DP-SGD}

While DP-SGD is the dominant approach for enforcing DP in medical deep learning, several studies explored alternative mechanisms that better match real-world deployment constraints, particularly when strict privacy budgets destabilize DP-SGD training or when data cannot be centralized. These alternatives fall into four recurring patterns (summarized in Table~\ref{tab:additional}).

\begin{table}[h]
    \centering
    \resizebox{\textwidth}{!}{%
        \begin{tabular}{p{3.5cm} >{\centering\arraybackslash}p{3.4cm} >{\centering\arraybackslash}p{5cm} >{\centering\arraybackslash}p{2.5cm}}
            \toprule
            \textbf{Study} & \textbf{Modalities} & \textbf{Downstream task}  & \textbf{Mech./account.} \\
            \midrule
            
            Al Aziz et al. \cite{31AZIZ2022104113} & Tabular data & Classification &  LM, EM/LC \\

            Almadhoun et al. \cite{76Inferenceattacksagainstdifferentiallyprivatequery} & Genomic data & Attack evaluation & LM/NA  \\

            Alsenani et al. \cite{23FAItH} & Sensor data & Classification &  LM, GM, EM/LC \\

            Ay et al. \cite{9Dp-FedAvg} & Brain MRI & Classification & NA/NA  \\

            Chen et al. \cite{16DP-FLMD} & DNA-sequence & Pattern discovery &  RR/NA \\

            D{\'\i}az et al. \cite{85diaz2025metric} & Brain MRI & Classification & GM/NA \\

            Faisal et al. \cite{52GeneratingPrivacyPreservingSyntheticMedicalData} & Chest X-ray & Classification &  GM/RDP \\

            Fan et al. \cite{82DP-MLD} & EEG, sensor data & Classification &  LM/LC \\

            Fang et al. \cite{83DP-CTGAN} & Tabular data & Tabular data generation &  GM/MA \\

            Fedeli et al. \cite{61fedeli2021privacy} & Tabular data & Regression & LM/NA  \\

            Gong et al. \cite{27gong2023federated} & Time series & Classification & GM/MA  \\

            Gwon et al. \cite{42GWON2024107738} & EHR & Data release &  RR, LM/NA \\

            Hatamizadeh et al. \cite{34tang2024personalized} & Tabular data & Classification &  NA/NA \\

            He et al. \cite{12DP-SL-GAN} & RNA-sequence & Classification &  GM/LC \\

            He et al. \cite{75Achievingdifferentialprivacyofgenomicdata} & Genomic data & Tabular data generation & LM/NA  \\

            Honkela et al. \cite{38honkela2018efficient} & RNA-sequence & Regression &  LM/LC \\

            Kim et al. \cite{77Privacy-preservingaggregationofpersonalhealthdatastreams} &  Sensor data & Classification & LM/LC  \\

            Kong et al. \cite{4FACLCancer} & Histopathology images & Classification & GM/NA  \\

            Li et al. \cite{11ADDETECTOR} & Pathological speech & Classification &  LM, GM/NA \\

            Li et al. \cite{44LI2019103138} & EHR & Classification  &  LM/NA \\

            Ming Y et al. \cite{40LU2022102298} & Histopathology images & Classification, regression &  GM/NA \\

            Movahedi et al. \cite{58EvaluatingClassifiersTrainedonDifferentiallyPrivateSyntheticHealthData} & Tabular data & Classification  & LM, EM/NA  \\

            Niinimäki et al. \cite{68Representationtransferfordifferentiallyprivatedrugsensitivityprediction} & Genomic data & Classification, regression &  LM/LC \\

            Sun et al. \cite{39SUN2023104404} &  Tabular data & Tabular data generation & GM/RDP  \\

            Tchouka et al. \cite{17NLPICD-10} & Unstructured clinical text & Classification &  LM, EM/NA\\ 

            Tram\`{e}r et al. \cite{80DifferentialPrivacywithBoundedPriors} & Genomic data & Feature selection &  LM, EM/NA \\

            Veeraragavan et al. \cite{81raghavan2024differentially} & EHR & Regression &  LM/LC \\

            Wang et al. \cite{60DifferentialPrivateFederatedTransferLearningforMentalHealthMonitoringinEverydaySettings} & Sensor data & Classification &  LM/NA\\

            Wei et al. \cite{29wei2024defedhdp} & Tabular data & Classification &  GM/LC \\

            Wu et al. \cite{79wu2021privacy} & Tabular data & Data release & RR/NA  \\

            Yuan et al. \cite{49yuan2019collaborative} & Chest X-ray & Classification &  GM/GDP \\

            Zhang et al. \cite{62BalancingCentralizedandLocalDifferentialPrivacyinPneumoniaDiagnosis} & Chest X-ray & Classification & GM/NA  \\

            Zhang et al. \cite{78zhang2022differential} & EHR & Data release & LM/LC  \\
            
            \bottomrule
        \end{tabular}%
    }
    \caption{Summary of studies applying alternative DP mechanisms beyond DP-SGD. Each entry includes the data modality, downstream task, and the DP mechanism and accounting method used. These include applications of local DP, encryption-based aggregation, federated learning (FL) variants, and generative modeling frameworks across a range of clinical modalities and tasks. AC: advanced composition; AD: Alzheimer's disease; EHR: electronic health record; EM: exponential mechanism; GDP: Gaussian differential privacy; GM: Gaussian mechanism; PD: Parkinson’s disease; NA: information not available; LC: linear composition; LM: Laplace mechanism; MA: moments accountant; PLD: privacy loss distribution; RDP: Rényi DP accountant; RR: randomized response.}
        \label{tab:additional}
\end{table}

The first pattern replaces gradient perturbation with privacy-preserving knowledge transfer. In the private aggregation of teacher ensembles (PATE) framework~\cite{PATE}, sensitive training data are partitioned across disjoint teacher models,
\begin{equation}
D = \bigcup_{k=1}^{K} D_k, \qquad D_i \cap D_j = \emptyset \text{ for } i \neq j
\end{equation}
and each teacher produces a class prediction. Votes for each class are tallied,
\begin{equation}
v_c(x) = \sum_{k=1}^{K} \mathbb{I}\big[T_k(x)=c\big],
\end{equation}
after which Laplace noise is added to the vote counts,
\begin{equation}
\tilde{v}_c(x) = v_c(x) + \mathrm{Lap}(1/\lambda),
\end{equation}
and the final label exposed to the student model is the noisy majority vote,
\begin{equation}
\hat{y}(x) = \arg\max_c \tilde{v}_c(x).
\end{equation}
By training only on noisy labels, student models never interact with private data. Studies using PATE, particularly for GAN-based synthetic data~\cite{PATEvsDPSGDGAN}, reported better utility at small $\epsilon$ than DP-SGD when data could be partitioned.

A second group of studies applied DP in federated or decentralized settings, enabling collaborative training without sharing data. DP noise was injected either locally at each client or centrally after aggregation. Applications included federated clustering of sensor data~\cite{23FAItH}, intensive care unit (ICU) monitoring~\cite{27gong2023federated}, pneumonia classification from chest X-rays~\cite{49yuan2019collaborative}, and MRI models using complex-valued networks~\cite{73Complex-ValuedFederatedLearningwithDifferentialPrivacy}. However, some papers (e.g.,~\cite{60DifferentialPrivateFederatedTransferLearningforMentalHealthMonitoringinEverydaySettings, 9Dp-FedAvg}) did not report $(\epsilon,\delta)$ or the privacy accountant, limiting reproducibility.

A third pattern perturbed data or outputs directly rather than modifying gradients. Laplace and exponential mechanisms were used for speech classification~\cite{11ADDETECTOR}, EHR and clinical text modeling~\cite{17NLPICD-10,31AZIZ2022104113}, and survival prediction~\cite{61fedeli2021privacy}. Local differential privacy was applied in tabular and EHR sharing~\cite{79wu2021privacy,42GWON2024107738}, avoiding composition accounting but typically reducing accuracy.

A fourth direction generated DP-synthetic data, decoupling downstream model training from private records. GANs and diffusion models were trained under DP constraints for RNA-sequence~\cite{12DP-SL-GAN}, tabular EHR~\cite{78zhang2022differential}, and CTGAN-based tabular data~\cite{83DP-CTGAN}. Sun et al.~\cite{39SUN2023104404} added noise to generator loss gradients rather than per-sample gradients, improving stability at low privacy budgets.

Across these studies, we observed substantial heterogeneity in reporting: more than one-third omitted essential details such as $(\epsilon,\delta)$ or the accountant used. Hybrid designs combining encryption and DP~\cite{DPMPC,DPplusMPC} illustrate that DP is increasingly implemented as a system-level deployment strategy, not only a gradient-level mechanism.

\subsection*{Defenses against attacks}

As privacy-preserving DL becomes increasingly integrated into clinical pipelines, adversarial attacks against these models have grown in sophistication. DP provides a formal framework for limiting such risks, but the real-world effectiveness of DP depends on implementation details, perturbation granularity, and the threat model assumed. Table~\ref{tab:attacks} summarizes empirical evaluations of DP-enabled defenses across various attack types in healthcare applications.

\begin{table}[h]
    \centering
    \resizebox{\textwidth}{!}{%
        \begin{tabular}{p{3.5cm} >{\centering\arraybackslash}p{3.4cm} >{\centering\arraybackslash}p{5cm} >{\centering\arraybackslash}p{3cm}}
            \toprule
            \textbf{Study} & \textbf{Perturbation} & \textbf{Attack type} & \textbf{Utilized DP} \\
            \midrule
            Almadhoun et al. \cite{76Inferenceattacksagainstdifferentiallyprivatequery} & GP & Attribute inference attack, MIA & DP-SGD\\
            Almadhoun et al. \cite{76Inferenceattacksagainstdifferentiallyprivatequery} & GP & MIA & DP-SGD\\
            Bingzhe et al. \cite{64wu2019p3sgd} & GP & Model-inversion attack &DP-SGD\\
            D{\'\i}az et al. \cite{85diaz2025metric} & Weight perturbation &  Client inference attack & Beyond DP-SGD\\
            DPFL \cite{3DPFRCovid19} & GP & MIA & DP-SGD \\
            Fan et al. \cite{26fan2023mitigating} & GP & MIA & DP-SGD \\
            Gwon et al. \cite{42GWON2024107738} & Input perturbation & MIA &  Beyond DP-SGD\\
            Hatamizadeh et al. \cite{33hatamizadeh2023gradient} & GP & GIA & DP-SGD \\
            Hyunwook et al. \cite{45kim2024synthetic} & Model-based perturbation & Re-identification risk assessment & Beyond DP-SGD\\
            Kaissis et al. \cite{66kaissis2021end} & GP & Model-inversion attack &DP-SGD\\
            Liangrui et al. \cite{63FedDP} & GP & GIA &DP-SGD\\
            Riess et al. \cite{73Complex-ValuedFederatedLearningwithDifferentialPrivacy} & GP & GIA & DP-SGD\\
            Shunrong et al. \cite{50DPFedSAM-Meas} & GP & MIA & DP-SGD\\
            Sun et al. \cite{39SUN2023104404} & GP & Identity disclosure attack & DP-SGD\\
            Sun et al. \cite{39SUN2023104404} & GP & Attribute disclosure attack & DP-SGD\\
            Tram\`{e}r et al. \cite{80DifferentialPrivacywithBoundedPriors} & Output perturbation & MIA & Beyond DP-SGD\\
            Veeraragavan et al. \cite{81raghavan2024differentially} & GP & MIA & DP-SGD\\
            You et al. \cite{23FAItH} & Output perturbation & MIA & Beyond DP-SGD \\
            Ziller et al. \cite{67ziller2024reconciling} & GP & Reconstruction attack & DP-SGD\\
            \bottomrule
        \end{tabular}%
    }
    \caption{Summary of attack evaluations on DP models in healthcare. Each row indicates the perturbation method used (e.g., gradient or input perturbation), the type of attack evaluated (e.g., membership inference, gradient inversion), and whether the study used DP-SGD or alternative DP methods. GIA: gradient inversion attack; GP: gradient perturbation; IP: input perturbation; MIA: membership inference attack.}
        \label{tab:attacks}
\end{table}

Membership inference attacks (MIAs) attempt to infer whether a specific sample was part of the model’s training dataset, a particularly serious risk in healthcare, where inclusion in a disease cohort may itself reveal sensitive clinical information. Multiple studies applied DP-SGD to protect against MIAs in classification tasks using imaging, sensor, and tabular data \cite{3DPFRCovid19, 26fan2023mitigating, 50DPFedSAM-Meas, 81raghavan2024differentially}. These defenses typically introduced Gaussian noise into gradient updates, reducing attack success rates even at moderate $\epsilon$ values.
However, the strength of protection varied across settings. Fan et al.\ \cite{26fan2023mitigating} noted that class imbalance and shallow architectures increased MIA vulnerability under DP, emphasizing the need to consider model and data characteristics when tuning privacy parameters. Almadhoun et al.\ \cite{76Inferenceattacksagainstdifferentiallyprivatequery} confirmed that even with DP-SGD, certain MIAs could still succeed under weak noise or limited clipping, highlighting the importance of robust DP configurations.

Beyond MIAs, several studies evaluated inversion-based threats, where attackers attempt to reconstruct sensitive inputs, such as medical images or EHR records, by analyzing shared gradients. These attacks are especially relevant in federated learning, where gradient exchange is common. For example, Hatamizadeh et al.\ \cite{33hatamizadeh2023gradient} and Liangrui et al.\ \cite{63FedDP} demonstrated that DP-SGD defenses could substantially degrade inversion quality but not eliminate leakage entirely. Ziller et al.\ \cite{67ziller2024reconciling} observed similar results in reconstruction attacks on multi-institutional imaging models.
Alternative perturbation schemes, such as weight noise \cite{85diaz2025metric} or complex-valued gradients \cite{73Complex-ValuedFederatedLearningwithDifferentialPrivacy}, were also explored as mitigation strategies. Still, these methods remain under-tested and lack standardized evaluation criteria, especially under stronger adversary assumptions (e.g., white-box gradient access).

LDP, which perturbs data before model access, was investigated in a few studies (e.g., \cite{23FAItH, 42GWON2024107738, 80DifferentialPrivacywithBoundedPriors}). While theoretically strong, LDP was shown to be insufficient in practice against sophisticated attacks, including attribute inference and partial MIAs. These findings underscore that DP defenses must often extend beyond input perturbation, especially in distributed learning settings.

Finally, a recurring recommendation across studies is the use of hybrid defenses, combining DP with cryptographic methods like secure aggregation or multiparty computation \cite{23FAItH, 79wu2021privacy, MPC, SMPC, DPMPC, DPplusMPC}. These layered strategies offer stronger guarantees but introduce practical constraints, including communication overhead and increased system complexity.

\section*{Discussion}

This scoping review synthesizes recent advances in applying DP to medical DL, with a particular focus on subgroup fairness, attack resilience, utility tradeoffs, and alternative privacy mechanisms. Across 74 studies, we observed growing maturity in technical implementations, but uneven adoption of standardized evaluation protocols and fairness-aware design practices.

Despite the theoretical strengths of DP \cite{dwork2014algorithmic}, practical implementations face well-documented tradeoffs. Performance consistently degrades under stricter privacy budgets, particularly at $\epsilon \approx 1$, where noise injection frequently leads to convergence failure or sharp accuracy declines \cite{22tayebi2024preserving, 53Adifferentialprivacybasedprototypicalnetworkformedicaldatalearning, 72arasteh2024differentialprivacyenablesfair}. However, several studies demonstrated that with strong pretraining \cite{22tayebi2024preserving, 47mehmood2024balancing}, stable architectures \cite{13TayebiDomainTransfer, 22tayebi2024preserving, 88FairandPrivateCTContrastAgentDetection}, and careful normalization \cite{72arasteh2024differentialprivacyenablesfair}, models can maintain clinically acceptable utility even under strong privacy guarantees ($\epsilon \approx $10). Imaging tasks, particularly chest X-ray and CT classification, showed resilience to DP noise \cite{13TayebiDomainTransfer, 22tayebi2024preserving}, while domains such as histopathology \cite{36adnan2022federated}, dermoscopy \cite{53Adifferentialprivacybasedprototypicalnetworkformedicaldatalearning}, and pathological speech \cite{72arasteh2024differentialprivacyenablesfair} proved more vulnerable, largely due to smaller datasets or higher input variability.

Beyond performance considerations, our findings highlight that DP introduces constraints that persist after training and influence real-world deployment \cite{cormode2018privacy, busch2025privacy, 66kaissis2021end}. Once a model is trained with DP, privacy guarantees remain tied to how often the model, or the underlying dataset, is accessed. Multiple users querying a DP-trained model, or repeated inference on the same data (e.g., auditing, calibration, federated reuse), incrementally consume the privacy budget through composition of privacy loss \cite{dwork2014algorithmic}. In adversarial scenarios, a malicious user could issue strategically engineered repeated queries to partially cancel injected noise or accelerate privacy depletion. As a result, DP must be treated as a finite, depletable resource over the model's lifecycle rather than a one-time property of training \cite{66kaissis2021end, 67ziller2024reconciling}. Operational deployments therefore require governance mechanisms, such as per-user privacy budgeting, access authentication, query logging, audit trails, and automatic rejection of further queries once the remaining privacy budget is exhausted \cite{cormode2018privacy, abowd2018us, near2025guidelines}. These constraints position DP not only as a mathematical guarantee but as a system-level design requirement that influences how DP models are shared, reused, and exposed in hospital or federated environments.

The privacy parameters $\epsilon$ and $\delta$ determine the strength of a DP guarantee, yet their practical implications are often under‐reported. A smaller $\epsilon$ corresponds to stronger privacy, meaning the model’s outputs are less sensitive to any single training example, but achieving low $\epsilon$ requires adding more noise, which reduces model performance. Across the reviewed studies, strict budgets ($\epsilon \approx 1$) frequently led to convergence failure or sharp utility drops, whereas moderate budgets ($\epsilon \approx 10$) preserved clinical accuracy in structured imaging tasks. The $\delta$ parameter, typically set to a negligible value (e.g., $1/N$ where $N$ is dataset size), represents the probability that the privacy guarantee may not hold; however, many papers report $\epsilon$ without disclosing $\delta$ or the privacy accountant used. These observations show that meaningful interpretation of privacy-utility tradeoffs requires reporting both parameters and contextualizing them in terms of clinical task fidelity rather than treating $\epsilon$ in isolation.

Fairness remains underexplored relative to utility. Only a subset of studies explicitly audit subgroup disparities under DP \cite{13TayebiDomainTransfer, 22tayebi2024preserving, 72arasteh2024differentialprivacyenablesfair, 88FairandPrivateCTContrastAgentDetection, 74chin2020generation}. These works focus mainly on classification and binary protected attributes (e.g., age, sex, race). Noise introduced by DP can interact with data imbalance and amplify disparities, particularly for underrepresented groups or tasks already exhibiting poor performance \cite{cummings2019compatibility, tran2021decision}. For example, Tayebi Arasteh et al.\ showed that DP disproportionately degraded performance for certain age groups in speech classification \cite{72arasteh2024differentialprivacyenablesfair}. However, a recent multi-institutional evaluation found that performance degradation under DP was uncorrelated with subgroup size and instead depended on task difficulty, with clinically complex or rare conditions affected most; at moderate privacy levels ($\epsilon \approx 5$–10), fairness remained stable, and some underrepresented groups even benefited from stronger privacy guarantees \cite{22tayebi2024preserving}. These results echo broader fairness literature showing that inequity often stems from dataset and labeling structure \cite{Mehrabi2021Survey, obermeyer2019dissecting}.
Despite these insights, fairness is rarely a design objective. Nearly all studies report disparities only post-hoc and at a single $\epsilon$ value, limiting understanding of privacy–fairness tradeoffs. Intersectional fairness (e.g., older female patients) and non-classification tasks such as regression and survival prediction remain largely unexamined \cite{26fan2023mitigating, 81raghavan2024differentially}. Fairness-aware optimization strategies, such as reweighting, constraint-based objectives, or adversarial debiasing, are virtually absent, despite maturity in the broader AI fairness community \cite{Mehrabi2021Survey}. Metrics such as subgroup AUC, underdiagnosis rate, and statistical parity difference have been used \cite{13TayebiDomainTransfer, 88FairandPrivateCTContrastAgentDetection, 74chin2020generation}, but no consensus exists on which metrics are most clinically meaningful or how to interpret them under noisy predictions.
To prevent fairness regressions, future DP pipelines should report subgroup performance across multiple $\epsilon$ levels, justify protected attribute selection based on clinical relevance, monitor fairness during deployment, and treat fairness as a first-class objective rather than an after-the-fact audit \cite{Mehrabi2021Survey, obermeyer2019dissecting}.

We also reviewed the threat landscape for privacy-preserving models. Empirical evaluations show that DP can substantially mitigate attacks such as membership inference and gradient inversion \cite{3DPFRCovid19, 26fan2023mitigating, 33hatamizadeh2023gradient, 67ziller2024reconciling}, but protections are highly context-dependent. Fan et al.\ \cite{26fan2023mitigating} emphasized that class imbalance and shallow architectures increased vulnerability. Almadhoun et al.\ \cite{76Inferenceattacksagainstdifferentiallyprivatequery} demonstrated residual attack success under weak noise, even with DP-SGD. Some studies also evaluated reconstruction attacks in FL \cite{63FedDP, truhn2024encrypted}, identity and attribute disclosure risks \cite{39SUN2023104404}, and defenses using weight perturbation \cite{85diaz2025metric} or complex-valued networks \cite{73Complex-ValuedFederatedLearningwithDifferentialPrivacy}. Importantly, LDP alone was shown to be insufficient under adversarial conditions \cite{42GWON2024107738, 80DifferentialPrivacywithBoundedPriors}.
Despite growing interest in attack resilience, several gaps remain. First, the majority of studies focus on a narrow set of attacks, most commonly MIAs, while overlooking others like training-time poisoning or client-level inference in federated setups. Second, many reports lack standardized threat models or omit key experimental details (e.g., batch size, gradient access scope), limiting reproducibility. Third, defenses are rarely evaluated under varying $\epsilon$ levels, making it difficult to assess tradeoffs between privacy strength and robustness.

This review has several limitations. First, while our search covered major venues and was supplemented by manual inclusion of key preprints, the rapid pace of publication means that very recent studies may not have been captured. Second, although we systematically included all eligible work applying DP to medical DL, the evidence base remains uneven: some clinical areas, such as genomics \cite{25zhang2022secure, 68Representationtransferfordifferentiallyprivatedrugsensitivityprediction} and survival modeling \cite{26fan2023mitigating, 81raghavan2024differentially}, are still underrepresented in the literature. Third, the overwhelming majority of empirical studies use DP-SGD; therefore, most quantitative synthesis naturally focuses on this mechanism, while alternative approaches (e.g., local DP, PATE, DP-synthetic data generation) are summarized conceptually due to limited and heterogeneous reporting. This reflects a limitation of the current field rather than a methodological bias of the review \cite{de2022unlocking, chua2024private}. Finally, our synthesis centers on technical and performance dimensions; perspectives from clinicians, patients, and data custodians were outside the scope but remain essential for translating privacy-preserving models into practice.

Several priorities emerge from this synthesis. First, given the systematic fairness gaps observed across studies, future DP work should include fairness evaluation across multiple $\epsilon$ levels \cite{Mehrabi2021Survey, obermeyer2019dissecting}. Second, benchmark datasets with demographic annotations and support for intersectional subgrouping are needed to enable comparable fairness audits under DP constraints \cite{mitchell2019model, gebru2021datasheets}. Third, future DP methods should integrate fairness during training rather than treating it as an after-the-fact audit; emerging techniques such as group-aware clipping or adaptive noise allocation offer promising directions \cite{cummings2019compatibility, tran2021decision}. Fourth, reproducibility requires clearer reporting: studies should disclose $(\epsilon,\delta)$, the accountant type, clipping norm, and performance at multiple privacy levels. Without such reporting, privacy–utility tradeoffs cannot be interpreted or compared \cite{dwork2014algorithmic, opacus_neurips2021}.
DP offers a principled mechanism to limit individual information leakage, but privacy alone does not guarantee equitable or reliable deployment. The evidence reviewed here shows that model utility, fairness, and attack resilience depend not only on privacy budgets but also on architectural choices, pretraining, dataset structure, and evaluation practices. Moving forward, integrated design, optimizing privacy, fairness, and utility together rather than treating them as isolated objectives, will be necessary for trustworthy clinical AI \cite{tabassi2023artificial, act2024eu}.

Based on the evidence reviewed, deploying DP in clinical AI requires coordinated technical and governance decisions. For developers, pretraining and compact CNN architectures (e.g., ResNet-9/18) repeatedly appeared in studies that maintained good utility under DP. Several works used group normalization to avoid batch-statistics leakage, though current evidence does not establish a universal best practice across modalities. Fairness auditing should become standard, with subgroup reporting across multiple privacy budgets and justification of protected attributes. For healthcare institutions, privacy budget should be treated as a depletable resource after deployment, requiring governance controls such as authenticated access, per-user query limits, logging, and audit trails. For policymakers and regulators, transparent reporting of $(\epsilon,\delta)$, the accountant used, clipping norm, and performance across privacy levels should be required to ensure comparability across systems \cite{66kaissis2021end, cummings2018role, near2025guidelines}. These practices move DP from a theoretical guarantee to an operational framework for safe, reproducible, and equitable deployment in clinical environments.


\section*{Methods}

This scoping review was conducted in accordance with the Preferred Reporting Items for Systematic Reviews and Meta-Analyses extension for Scoping Reviews (PRISMA-ScR) guidelines~\cite{tricco2018prisma, page2021prisma} to ensure methodological rigor and transparency \cite{buess2025large}. The complete PRISMA-ScR checklist is provided in Supplementary Table 1. The methodological approach to data collection, article selection, and results synthesis is described in detail below.

Any study was eligible if it was published on or before March 1, 2025, and explicitly focused on empirical applications of DP in conjunction with DL methods for healthcare-related tasks. Eligible articles included original peer-reviewed journal papers, conference proceedings, and impactful preprints in English. Studies purely theoretical in nature or without explicit medical or clinical applications were excluded. Review articles, editorials, and commentaries were also excluded to focus exclusively on primary empirical contributions. 

\subsection*{Information sources}
We systematically searched four electronic databases: PubMed, IEEE Xplore, ACM Digital Library, and Web of Science. Database queries were performed on March 1, 2025, and search results were imported into Rayyan software~\cite{ouzzani2016rayyan}, a tool designed to facilitate the systematic screening of literature and duplicate removal. A detailed description of the database-specific search queries is provided in Supplementary Table 2.

\subsection*{Search strategy}
The search strategy was carefully designed to identify publications at the intersection of DP, DL, and medical applications. Specifically, the search strings combined three primary keyword groups: (i) differential privacy techniques (e.g., ``differential privacy,'' ``differentially private,'' ``DP-SGD''), (ii) deep learning and related machine learning methodologies (e.g., ``deep learning,'' ``machine learning,'' ``neural network,'' ``artificial intelligence''), and (iii) healthcare and clinical applications (e.g., ``medical,'' ``clinical,'' ``healthcare,'' ``biomedical,'' ``EHR,'' ``imaging,'' ``radiology,'' ``text,'' ``speech''). 

Each database query was adapted to database-specific indexing conventions to maximize search sensitivity. The comprehensive queries and exact search strings used, along with the number of articles retrieved from each database, are summarized in Supplementary Table 2. Additionally, manual searches were conducted by screening reference lists from key articles and reviewing recent preprints and related studies to capture any relevant works missed by the database searches.

\subsection*{Inclusion and exclusion criteria}
The literature selection involved multiple screening steps: (i) Initial screening included automated duplicate removal and manual title and abstract screening in Rayyan \cite{ouzzani2016rayyan}. Articles were excluded if they lacked relevance to DP or healthcare applications. (ii) Full-text reviews were conducted on the remaining articles to ensure alignment with inclusion criteria, specifically the empirical application of DP with DL techniques in medical contexts. Articles with purely theoretical content, lacking empirical evaluation, or applied to non-healthcare domains were excluded. The selection process ensured proportional representation across different medical application domains, including medical imaging, electronic health records, predictive analytics, and personalized medicine.

\section*{Data availability}
No datasets were generated or analyzed during the current study. This is a scoping review based entirely on previously published literature. 

\section*{Code availability}
No custom code was developed or used in this study. All analyses were conducted through manual review of existing publications.

\section*{Acknowledgements}
We acknowledge financial support by Deutsche Forschungsgemeinschaft. DT was supported by grants from the DFG (NE 2136/3-1, LI3893/6-1, TR 1700/7-1) and is supported by the German Federal Ministry of Education (TRANSFORM LIVER, 031L0312A; SWAG, 01KD2215B) and the European Union’s Horizon Europe and innovation programme (ODELIA [Open Consortium for Decentralized Medical Artificial Intelligence], 101057091). The funders played no role in the design or execution of the study. The authors of this work take full responsibility for its content.

\section*{Author contributions}
The formal analysis was conducted by MM and STA. The original draft was written by MM and STA. The literature search was performed by STA, and the screening was performed by MM, MV, ML, and STA. MR, DT, and STA provided clinical expertise. MM, MV, ML, MR, DT, AM, and STA provided technical expertise. STA defined the study. Figures were designed by MM. All authors read the manuscript, contributed to the editing, and agreed to the submission of this paper.

\section*{Competing interests}
ML is employed by Generali Deutschland Services GmbH, Germany. DT received honoraria for lectures by Bayer, GE, Roche, AstraZeneca, and Philips and holds shares in StratifAI GmbH, Germany, and in Synagen GmbH, Germany. AM is an associate editor at IEEE Transactions on Medical Imaging. STA is an editorial board at Communications Medicine and at European Radiology Experimental, a trainee editorial board at Radiology: Artificial Intelligence, and is partially employed by Synagen GmbH, Germany. The other authors do not have any competing interests to disclose.

\section*{Ethics statement}
No human or animal subjects are involved in this study.

\newpage

\noindent\textbf{List of Abbreviations}
\begin{itemize}
    \item AD - Alzheimer's disease
    \item AI - Artificial intelligence
    \item AUC - Area under the receiver operating characteristic curve
    \item BN - Batch normalization
    \item CT - Computed tomography
    \item DL - Deep learning
    \item DP - Differential privacy
    \item DP-SGD - Differentially-private stochastic gradient descent
    \item ECG - Electrocardiogram
    \item EEG - Electroencephalography
    \item EHR - Electronic health record
    \item EM - Exponential mechanism
    \item FDR - False discovery rate
    \item FID - Fréchet inception distance
    \item FL - Federated learning
    \item FOR - False omission rate
    \item FPR - False positive rate
    \item GAN - Generative adversarial network
    \item GIA - Gradient inversion attack
    \item GM - Gaussian mechanism
    \item GN - Group normalization
    \item LM - Laplace mechanism
    \item LDP - Local differential privacy
    \item MA - Moments accountant
    \item MIA - Membership inference attack
    \item MRI - Magnetic resonance imaging
    \item MSN - Mode-specific normalization
    \item PATE - Private aggregation of teacher ensembles
    \item PD - Parkinson’s disease
    \item PRISMA - Preferred reporting items for systematic reviews and meta-analyses
    \item PRISMA-ScR - PRISMA extension for scoping reviews
    \item PtD - Statistical parity difference
    \item RDP - Rényi differential privacy
    \item SGD - Stochastic gradient descent
    \item TPR - True positive rate
    \item ViT - Vision transformer
\end{itemize}

\newpage
\bibliography{sn-bibliography}

\newpage
\appendix
\renewcommand{\thetable}{S.\arabic{table}}
\setcounter{table}{0} 

\section{Supplementary information}

\begin{longtable}{p{3cm} p{1cm} p{6cm} p{1.5cm}}
\toprule
\textbf{Section} & \textbf{Item} & \textbf{PRISMA-ScR checklist item} & \textbf{Status} \\
\midrule
\endfirsthead

\toprule
\textbf{Section} & \textbf{Item} & \textbf{PRISMA-ScR checklist item} & \textbf{Status} \\
\midrule
\endhead

\endfoot

\textbf{Title} &  &  & \\
\addlinespace
\hline
\addlinespace
Title & 1 & Identify the report as a scoping review. & \checkmark \\
\addlinespace
\hline
\addlinespace
\textbf{Abstract} &  &  & \\
\addlinespace
\hline
\addlinespace
Structured summary & 2 & Provide a structured summary that includes (as applicable): background, objectives, eligibility criteria, sources of evidence, charting methods, results, and conclusions that relate to the review questions and objectives. & \checkmark \\
\addlinespace
\hline
\addlinespace
\textbf{Introduction} &  &  & \\
\addlinespace
\hline
\addlinespace
Rationale & 3 & Describe the rationale for the review in the context of what is already known. Explain why the review questions/objectives lend themselves to a scoping review approach. & \checkmark \\
\addlinespace
Objectives & 4 & Provide an explicit statement of the questions and objectives being addressed with reference to their key elements (e.g., population or participants, concepts, and context) or other relevant key elements used to conceptualize the review questions and/or objectives. & \checkmark \\
\addlinespace
\hline
\addlinespace
\textbf{Methods} &  &  & \\
\addlinespace
\hline
\addlinespace
Protocol and registration & 5 & Indicate whether a review protocol exists; state if and where it can be accessed (e.g., a Web address); and if available, provide registration information, including the registration number. & \checkmark \\
\addlinespace
Eligibility criteria & 6 & Specify characteristics of the sources of evidence used as eligibility criteria (e.g., years considered, language, and publication status), and provide a rationale. & \checkmark \\
\addlinespace
Information sources & 7 & Describe all information sources in the search (e.g., databases with dates of coverage and contact with authors to identify additional sources), as well as the date the most recent search was executed. & \checkmark \\
\addlinespace
Search & 8 & Present the full electronic search strategy for at least 1 database, including any limits used, such that it could be repeated. & \checkmark \\
\addlinespace
Selection of sources of evidence & 9 & State the process for selecting sources of evidence (i.e., screening and eligibility) included in the scoping review. & \checkmark \\
\addlinespace
Data charting process & 10 & Describe the methods of charting data from the included sources of evidence (e.g., calibrated forms or forms that have been tested by the team before their use, and whether data charting was done independently or in duplicate) and any processes for obtaining and confirming data from investigators. & \checkmark \\
\addlinespace
Data items & 11 & List and define all variables for which data were sought and any assumptions and simplifications made. & \checkmark \\
\addlinespace
Critical appraisal of individual sources of evidence & 12 & If done, provide a rationale for conducting a critical appraisal of included sources of evidence; describe the methods used and how this information was used in any data synthesis (if appropriate). & NA \\
\addlinespace
Synthesis of results & 13 & Describe the methods of handling and summarizing the data that were charted. & \checkmark \\
\addlinespace
\hline
\addlinespace
\textbf{Results} &  &  & \\
\addlinespace
\hline
\addlinespace
Selection of sources of evidence & 14 & Give numbers of sources of evidence screened, assessed for eligibility, and included in the review, with reasons for exclusions at each stage, ideally using a flow diagram. & \checkmark \\
\addlinespace
Characteristics of sources of evidence & 15 & For each source of evidence, present characteristics for which data were charted and provide the citations & \checkmark \\
\addlinespace
Critical appraisal within sources of evidence & 16 & If done, present data on critical appraisal of included sources of evidence (see item 12). & NA \\
\addlinespace
Results of individual sources of evidence & 17 & For each included source of evidence, present the relevant data that were charted that relate to the review questions and objectives. & \checkmark \\
\addlinespace
Synthesis of results & 18 & Summarize and/or present the charting results as they relate to the review questions and objectives & \checkmark \\
\addlinespace
\hline
\addlinespace
\textbf{Discussion} &  &  & \\
\addlinespace
\hline
\addlinespace
Summary of evidence & 19 & Summarize the main results (including an overview of concepts, themes, and types of evidence available), link to the review questions and objectives, and consider the relevance to key groups. & \checkmark \\
\addlinespace
Limitations & 20 & Discuss the limitations of the scoping review process & \checkmark \\
\addlinespace
Conclusions & 21 & Provide a general interpretation of the results w.r.t the review questions and objectives, as well as potential implications and/or next steps. & \checkmark \\
\addlinespace
\hline
\addlinespace
\textbf{Funding} &  &  & \\
\addlinespace
\hline
\addlinespace
Funding & 19 & Describe sources of funding for the included sources of evidence, as well as sources of funding for the scoping review. Describe the role of the funders of the scoping review. & \checkmark \\
\addlinespace

\bottomrule
\addlinespace
\caption{PRISMA-ScR checklist \cite{tricco2018prisma}} \\
\label{tab:checklist}
\end{longtable}

\begin{table}[h]
    \centering
    \resizebox{\textwidth}{!}{
    \begin{tabular}{p{2.9cm} p{1.5cm} p{9.5cm}}
        \toprule
        \textbf{Database} & \textbf{Results} & \textbf{Search string} \\
        \midrule
        PubMed & 63 & ("differential privacy"[TIAB] OR "differentially private"[TIAB] OR "DP-SGD"[TIAB]) AND ("deep learning"[TIAB] OR "machine learning"[TIAB] OR "neural network"[TIAB]) AND ("medicine"[TIAB] OR "medical"[TIAB] OR "clinical"[TIAB] OR "biomedical"[TIAB] OR "health"[TIAB] OR "EHR"[TIAB] OR "imaging"[TIAB] OR "radiology"[TIAB] OR "text"[TIAB] OR "speech"[TIAB]) NOT (review[PT]) \\
         & 84 & ("differential privacy" OR "differentially private" OR "DP-SGD") AND ("deep learning" OR "machine learning" OR "neural network" OR "artificial intelligence") AND ("medical" OR "medicine" OR "clinical" OR "biomedical" OR "healthcare" OR "EHR" OR "imaging" OR "radiology" OR "MRI" OR "CT" OR "ultrasound" OR "speech" OR "text" OR "video") NOT (review[PT]) \\
         & 417 & ( "differential privacy"[TIAB] OR "differentially private"[TIAB] OR "DP-SGD"[TIAB] OR "privacy preserving"[TIAB] OR "privacy-preserving"[TIAB] OR "securing"[TIAB] OR "safeguarding"[TIAB] ) AND ( "deep learning"[TIAB] OR "machine learning"[TIAB] OR "artificial intelligence"[TIAB] OR "neural network"[TIAB] ) AND ( "medical"[TIAB] OR "medicine"[TIAB] OR "clinical"[TIAB] OR "healthcare"[TIAB] OR "biomedical"[TIAB] OR "EHR"[TIAB] OR "electronic health record"[TIAB] OR "imaging"[TIAB] OR "radiology"[TIAB] OR "text"[TIAB] OR "speech"[TIAB] ) NOT (review[PT]) \\
        \hline
        \addlinespace
        IEEE Xplore & 676 & ("differential privacy" OR "differentially private" OR "DP-SGD") AND ("deep learning" OR "machine learning" OR "neural network") AND ("medicine" OR "medical" OR "clinical" OR "healthcare" OR "biomedical" OR "EHR" OR "imaging" OR "radiology" OR "text" OR "speech") \\
        \hline
        \addlinespace
        Web of Science & 494 & "differential privacy" AND ("deep learning" OR "machine learning") AND (medical OR medicine OR clinical OR healthcare OR biomedical OR imaging OR EHR OR radiology OR text OR speech) \\
         & 571 & ("differential privacy" OR "differentially private" OR "DP-SGD") AND ("deep learning" OR "machine learning" OR "artificial intelligence" OR "neural network") AND (medical OR medicine OR clinical OR healthcare OR biomedical OR imaging OR EHR OR radiology OR text OR speech) \\
        \hline
        \addlinespace
        ACM Digital Library & 999 & "differential privacy" AND ("deep learning") AND (medical OR EHR OR imaging) \\
         & 938 & "differential privacy" AND ("artificial intelligence") AND (medical OR clinical OR healthcare OR EHR OR imaging OR radiology OR biomedical) \\
        \bottomrule
    \end{tabular}
    }
    \caption{Search results from different databases}
    \label{tab:queries}
\end{table}

\end{document}